\documentclass[10pt,journal,letterpaper,compsoc]{IEEEtran}

\ifCLASSOPTIONcompsoc
\else
\fi

\usepackage[utf8x]{inputenc}
\usepackage{graphicx}
\usepackage{subfig}
\usepackage{amsmath}
\usepackage{amssymb}
\usepackage{multirow}
\usepackage{rotating}
\usepackage{fancyhdr}
\usepackage{latexsym}
\usepackage{url}
\usepackage{algorithm}
\usepackage{algorithmic}
\usepackage{bm}
\usepackage{stmaryrd}
\usepackage{dsfont}

\usepackage{xspace}
\makeatletter
\DeclareRobustCommand\onedot{\futurelet\@let@token\@onedot}
\def\@onedot{\ifx\@let@token.\else.\null\fi\xspace}
\def\eg{\emph{e.g}\onedot} 
\def\ie{\emph{i.e}\onedot}

\def\etal{\emph{et al}\onedot}
\makeatother

\newcommand{\eref}[1]{Eq.~(\ref{#1})}
\newcommand{\Fref}[1]{Figure~\ref{#1}}
\newcommand{\fref}[1]{Fig.~\ref{#1}}

\newcommand{\sref}[1]{Sec.~\ref{#1}}

\begin{document}

\makeatletter
\let\@oldmaketitle\@maketitle
\renewcommand{\@maketitle}{\@oldmaketitle
  \includegraphics[width=\linewidth,height=6\baselineskip]
    {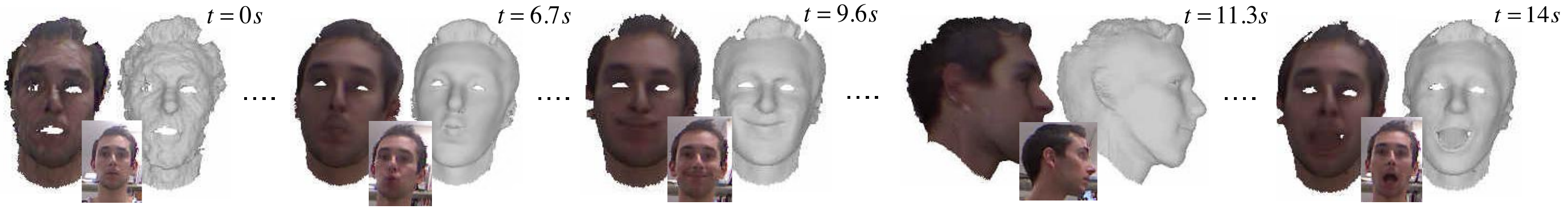}
    \captionof{figure}{Real-time simultaneous 3D head modeling and facial motion capture using an RGB-D camera. The 3D model of the head of a moving person refines over time (left to right) while the facial motion is being captured.  
    }
     \rule{\textwidth}{0.4pt}
    \label{fig:frontpage}
    \vspace{-0.4cm}
    }
\makeatother

%
\title{Real-time Simultaneous 3D Head Modeling and Facial Motion Capture with an RGB-D camera}
%
%
%
%

\author{Diego~Thomas 
\IEEEcompsocitemizethanks{\IEEEcompsocthanksitem D. Thomas is with the Kyushu University, Fukuoka, Japan.\protect\\ 
E-mail: thomas@ait.kyushu-u.ac.jp}
\thanks{}}

\IEEEcompsoctitleabstractindextext{%
\begin{abstract}

We propose a method to build in real-time animated 3D head models using a consumer-grade RGB-D camera. Our proposed method is the first one to provide simultaneously comprehensive facial motion tracking and a detailed 3D model of the user's head. Anyone's head can be instantly reconstructed and his facial motion captured without requiring any training or pre-scanning. The user starts facing the camera with a neutral expression in the first frame, but is free to move, talk and change his face expression as he wills otherwise. The facial motion is captured using a blendshape animation model while geometric details are captured using a Deviation image mapped over the template mesh. We contribute with an efficient algorithm to grow and refine the deforming 3D model of the head on-the-fly and in real-time. We demonstrate robust and high-fidelity simultaneous facial motion capture and 3D head modeling results on a wide range of subjects with various head poses and facial expressions. 

\end{abstract}

}

\maketitle


\IEEEdisplaynotcompsoctitleabstractindextext

%
\IEEEpeerreviewmaketitle

\section{Introduction}
Real-time dynamic 3D head reconstruction is a task of great importance and of extraordinary difficulty. The 3D shape of someones's head contains a lot of personal information that can be used for identification (like unlocking a smartphone) and communication (through an Augmented Reality platform for example) purpose. Unlike 2D color images, 3D models are invariant to environment changes such as illumination or make-up. Thus they are a powerful and reliable tool for security application. With the recent development and miniaturization of depth sensors, RGB-D cameras have become widely available to many consumers (for example, the new Apple iPhone generation is equipped with a frontal RGB-D camera), which also brings new challenges. One of these challenges, which motivates this work, is that users are not experts in 3D scanning: sometimes impatient (like children) or sometimes unable to follow instructions (like elderly people). Thus it is crucial to enable detailed 3D head reconstruction, while requiring few efforts (ideally none) for the user. In particular, this means that the user must be free to move and talk while to 3D model is being built. 

In the RGB-D Vision community, dynamic 3D scene reconstruction is a recent trend of research with many open problems, such as how to deal with fast motion, noise or occlusions. In this work, we push forward the research on dynamic 3D head reconstruction (\ie, simultaneous 3D head modeling and facial motion capture) using a single RGB-D camera.

Marker-less facial motion capture, on one hand, is well established in the computer graphics community. Several methods using template 3D models \cite{Bouaziz:2013, Hsieh:2015, Weise:2011} achieve real-time accurate facial motion capture from videos of RGB-D images. On the other hand, applications of dense 3D modeling techniques to build 3D head models from static scenes \cite{Pavan:2013, Hernandez:2012, Zollhofer:2011} showed compelling results in terms of details in the produced 3D models. Recently, an extension of the popular KinectFusion algorithm \cite{Newcombe:2011} called DynamicFusion \cite{Newcombe:2015} was proposed that can handle even dynamic scenes. 

While recent advances have shown compelling results in either facial motion capture or dense 3D modeling, they do not allow to produce both results at the same time. Though DynamicFusion \cite{Newcombe:2015} allows to capture deformations of the face, the results are limited compared to those obtained with facial motion capture systems (\eg, eyelids movements can not be captured). Moreover, the obtained deformations are not intuitive for animation purpose (animations such as "mouth open" or "mouth closed" are more intuitive to animate the face). This is because the head is animated using unstructured deformation nodes, without any semantic meaning. Note that DynamicFusion was designed for a more general purpose: dynamic scene reconstruction, while in this work we focus on dynamic 3D head modeling. 

We propose a new method to simultaneously build a high-fidelity 3D model of the head and capture its facial motion using an RGB-D camera (\fref{fig:frontpage}). To do so, (1) we introduce a new dynamic 3D representation of the head based on blendshapes \cite{Weise:2011} and Deviation images \cite{Diego:2013}, and (2) we propose an efficient method to fuse in real-time input RGB-D data into our proposed 3D representation. While blendshape coefficients encode facial expressions, the Deviation image augments the blendshape meshes to encode the geometric details of the user's head. The head position and its facial motion are tracked in real-time using a facial motion capture approach, while the 3D model of the head grows and refines on-the-fly with input RGB-D images using a running median strategy. Our proposed method do not require any training, fine fitting or pre-scanning to produce accurate animation results and highly detailed 3D models. Our main contribution is to propose the first algorithm that is able to build, in real-time, detailed (with Deviation and color images) and comprehensive (with blendshape representation) animated 3D models of the user's head. We remark that a part of this work has been reported in \cite{Thomas:CVPR}\footnote{The main changes compared with \cite{Thomas:CVPR} are (1) adding occlusion handling and (2) using running median instead of running average.}.

\begin{figure*}[t]
\begin{center}
	\includegraphics[width=1.0\linewidth]{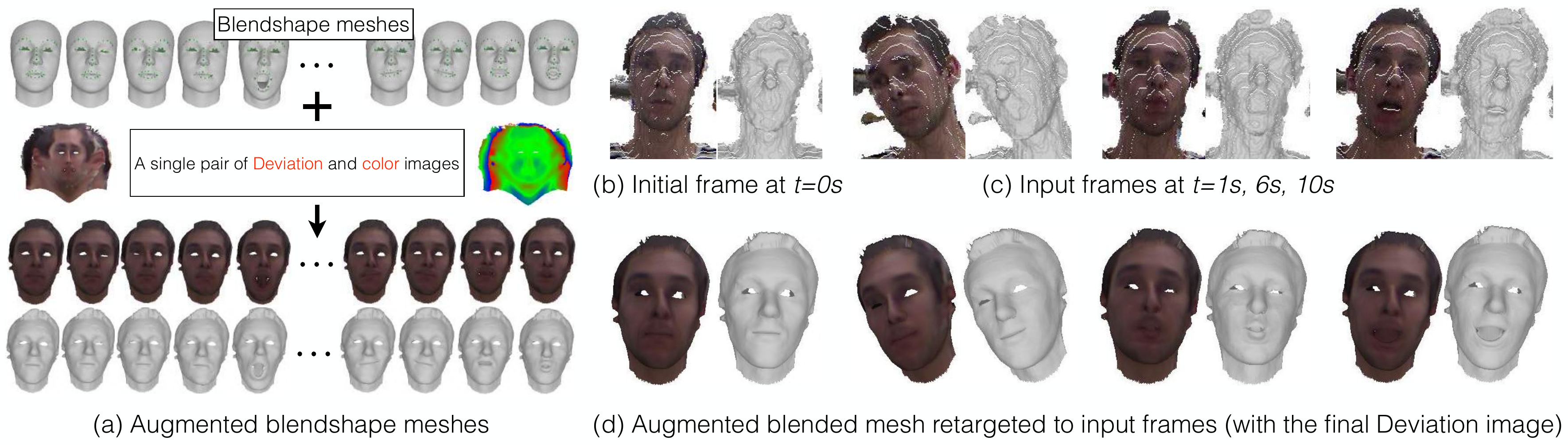}
\end{center}
   \caption{The main idea of our proposed method is that a single pair of Deviation and color images is sufficient to augment all template blendshape meshes and build a space of high fidelity user-specific facial expressions. While the blendshape meshes represent the general shape of the head and facial expressions, the Deviation image represents the fine details of the user's head.}
     \rule{\textwidth}{0.4pt}
\label{fig:concept}
\vspace{-0.4cm}
\end{figure*}

\section{Related works}

There are two categories of closely related work: 1) real-time facial motion capture and 2) real-time dense 3D reconstruction. While facial motion capture systems strive to capture high fidelity facial expressions, dense 3D reconstruction methods focus on constructing detailed 3D models of a target scene (the user's head in our case).

\subsection{Real-time facial motion capture}
Research on real-time marker-free facial motion capture using RGB-D sensors have raised much interest in computer graphics in the last few years \cite{Bouaziz:2013, Cao:2013, Chen:2013, Hsieh:2015, Li:2013, Weise:2011, Weise:2009}. The use of blendshapes introduced by Weise \etal in \cite{Weise:2011} for tracking facial motions has become popular and motivated many researchers to build more portable \cite{Cao:2013} or user-friendly systems \cite{Bouaziz:2013, Hsieh:2015, Li:2013}. In these works, facial expressions are expressed using a weighted sum of blendshapes. The tracking process then consists of (1) estimating the head pose and (2) optimizing the weights of each blendshape to fit the input RGB-D image. In \cite{Cao:2013, Weise:2011} the blendshapes were first fit to the user's face in a pre-processing training stage where the user was asked to perform several pre-defined expressions. Calibration-free systems were proposed in \cite{Bouaziz:2013, Hsieh:2015, Li:2013} where the neutral blendshape was adjusted on-the-fly to fit the input images. Sparse facial features were combined with depth data in \cite{Cao:2014, Cao:2014:2, Hsieh:2015, Li:2013} to improve tracking. Though compelling results were reported, much efforts were made on capturing high fidelity facial motions (for retargeting purpose), but the geometric details of the built 3D models were not as good as those obtained with state-of-the-art dense 3D modeling methods \cite{Newcombe:2011}.

Chen \etal \cite{Chen:2013} demonstrated that a template 3D model with geometric details close to the shape of the user's face can improve the facial motion tracking quality. In this work, the template mesh was built offline by scanning the user's face in a neutral expression. The template mesh was then incrementally deformed using embedded deformation \cite{Sumner:2007} to fit the input depth images. High fidelity facial motions were obtained but at the cost of a pre-processing scanning stage required to build the user-specific template mesh. Moreover, parts of the user's head that do not animate (\eg the ears or the hair) were simply ignored and not modelled.

\subsection{Real-time dense 3D reconstruction}
Low-cost depth cameras have spurred a flurry of research on real-time dense 3D reconstruction of indoor scenes. In KinectFusion, introduced by Newcombe \etal \cite{Newcombe:2011} and all follow-up research \cite{Henry:2013, NieBner:2013, Roth:2012, Whelan:2013, Whelan:2012, Zeng:2012}, the 3D model is represented as a volumetric Truncated Signed Distance Function (TSDF) \cite{Curless:1996}, and depth measurements of a static scene are fused into the TSDF to grow the 3D model. Applications of 3D reconstruction using RGB or RGB-D cameras to build a human avatar were proposed using either a single camera \cite{Pavan:2013, Hernandez:2012, Zollhofer:2011} or multiple cameras \cite{Kainz:2012}. The user was then assumed to hold still during the whole scanning period.

Recently, much interest has been given to reconstruct 3D models of dynamic scenes. Dou \etal \cite{Dou:2013} introduced a directional distance function to build dynamic 3D models of the human body offline. In \cite{Dou:2014}, static parts of the scene were pre-scanned offline, and movements of the body were tracked online. Zhang \etal \cite{Zhang:2014} proposed to merge different partial 3D scans obtained offline with KinectFusion in different poses into a single canonical 3D model. More recently, Newcombe \etal \cite{Newcombe:2015} extended KinectFusion to DynamicFusion, which allows capturing dynamic changes in the volumetric TSDF in real-time by using embedded deformation \cite{Sumner:2007}. Compelling results were reported for real-time dynamic 3D face modeling in terms of geometric accuracy. However, in terms of facial motion capture, the results were not as good as those reported in \cite{Cao:2013, Hsieh:2015}. To improve the restricted range of motions, extensions such as \cite{bodyfusion, volumedeform, killingfusion} were proposed. However, none of the above mentioned methods can achieve dynamic reconstruction of scenes that move quickly from a close to open topology. Therefore, the user must keep the mouth open for a few seconds at the beginning of the scanning process, which is not practical.

For the special case of 3D facial reconstruction, several methods that use a template mesh were proposed. In \cite{Zollhofer:2011, thies2015} a morphable face model was non-rigidly fitted to the depth data. However it is difficult and time consuming to accurately fit a dense 3D mesh to a sequence of depth images, which precludes from real-time application. Cao \etal \cite{cao2015real} and Garrido \etal \cite{Garrido:2016} proposed to combine blendshape meshes with machine learning to regress 3D wrinkle from color images of the face. However the reconstructed model is limited to the frontal face only. \cite{Cao2016, Ichim:2015}, on the other hand, proposed to reconstruct a full head 3D avatar of a user from multiple color images. However, this method does not work on-the-fly as it requires pre-processing using all captured images, and uses a computationally expensive non-rigid fitting algorithm. Moreover, even though the obtained template mesh approximates the shape of the user's 3D face, it does not accurately reflect the geometry.

We use a Deviation image \cite{Diego:2013} mapped over blendshapes because it is light in memory yet produces accurate 3D models. By using blendshapes we also achieve state-of-the-art facial motion tracking performances \cite{Hsieh:2015}.

\section{Proposed Dynamic 3D representation of the head}
\label{sec:model}
We introduce a new dynamic 3D representation of the head that allows us to capture facial motions as well as fine geometric details of the user's head. We propose to augment the blendshape meshes \cite{Weise:2011} with a Deviation image \cite{Diego:2013}. While blendshape coefficients encode facial expressions, the Deviation image encodes the geometric deviations of the user's head to the template mesh. We also build a color image for better visual impression.  

\subsection{Deviation mapping}
Our proposed 3D head representation is inspired by the displacement mapping (or bump mapping) widely used in Computer Graphics. The concept of Displacement mapping is to record in a separate texture image the distance from the surface to the textured 3D mesh in the direction of the normal vector. Then each point of the 3D mesh can be displaced along the local surface normal, according to the value at the corresponding texture coordinate, to obtain the detailed 3D geometry. Standard Displacement maps are built for a single 3D mesh. In our work we propose to build a single displacement image (that we call Deviation image) that is shared by all blendshape meshes (one mesh for each base expression). This allows us to have a dynamic 3D model of the head with detailed 3D geometry that is preserved in any expression. Note that in contrast to displacement mapping where the displacement is along the standard normal vector, our Deviation image records the displacement along the blended normal vector (this will be detailed below).

\subsection{Blendshape representation}
We briefly recall the blendshape representation, commonly used in facial motion capture systems \cite{Bouaziz:2013, Cao:2013, Hsieh:2015, Li:2013, Weise:2011}. Facial expressions are represented using a set of blendshape meshes $\{\mathbf{B}_i\}_{i \in [0:n]}$ ($n=27$ in our experiments), where $\mathbf{B}_0$ is the mesh with neutral expression and $\mathbf{B}_i$, $i > 0$ are the meshes in various base expressions. All blendshape meshes have the same number of vertices and share the same triangulation. A 3D point at the surface of the head is expressed as a linear combination of the blendshape meshes: $\mathbf{M}(x) = \mathbf{B}_0 + \sum\limits_{i=1}^n x_i \mathbf{\hat{B}}_i$, where $x = [x_1, x_2,  ..., x_{n}]$ are the blendshape coefficients (ranging from $0$ to $1$) and $\mathbf{\hat{B}}_i = \mathbf{B}_i - \mathbf{B}_0$ for $i \in [1:n]$. We call $\mathbf{M}(x)$ the blended mesh. 

The blendshape representation is an efficient way to accurately and quickly capture facial motions. However, because it is a template-based representation, it is not possible to capture the fine geometric details of the user's head (the hair for example can not be modeled). This is because real-time, accurate fitting of the template 3D meshes to input RGB-D images is a difficult task. Moreover, the resolution of the blendshape meshes is insufficient to capture fine geometric details. We overcome this limitation by augmenting the set of blendshape meshes with a single pair of Deviation and color images (as illustrated in \fref{fig:concept}).

\begin{figure}
\begin{center}
	\includegraphics[width=1.0\linewidth]{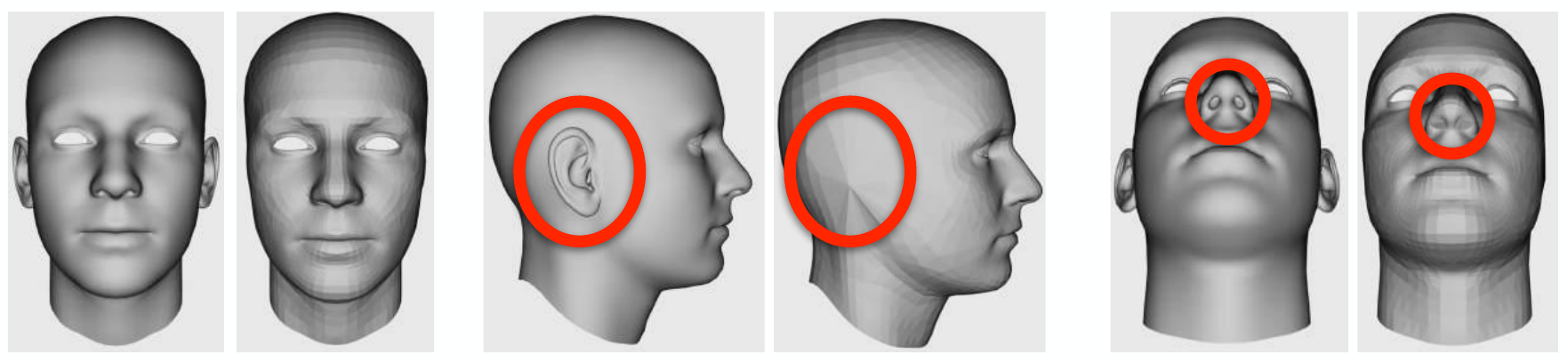}
\end{center}
   \caption{Original and modified blendshape mesh with neutral expression ($\mathbf{B}_0$). For each pose, the original mesh is on the left side and our modified mesh is on the right side. Modified areas are highlighted by the red circles.}
\label{fig:template}
\vspace{-0.4cm}
\end{figure}

We slightly simplify the original blendshape meshes \cite{Weise:2011} around the ears and around the nose. This is because these areas are too much detailed for our proposed 3D representation. We instead record geometric details of the head in the Deviation image. The original and modified templates with highlighted modified areas are shown in \fref{fig:template}.

\subsection{Augmented blendshapes}
Each vertex in the blendshape meshes has texture coordinates (the same vertex in different base expression has the same texture coordinates). We propose to build an additional texture image, called Deviation image that encodes the deviations of the user's head to the blendshape meshes in the direction of the blended normal vectors. Our proposed 3D representation is illustrated in \fref{fig:concept} (a) and detailed below.

In addition to the 3D positions of the vertices $\{ \{\mathbf{B}_i(j) \}_{j \in[0:l]} \}_{i \in [0:n]}$ in all blendshape meshes (where $l+1$ is the number of vertices), we also have the values of the normal vectors $\{ \{ \mathbf{Nmle}_i(j) \}_{j \in[0:l]} \}_{i \in [0:n]}$ and the list of triangular faces $\{ \mathbf{F}(j) = [s_0^j, s_1^j, s_2^j] \} _{j \in[0:f]}$, where $f+1$ is the number of faces and $[s_0^j, s_1^j, s_2^j]$ are the indices in $\{ \mathbf{B}_i \}_{i \in [0:n]}$ of the three vertices that are the summits of the $j^{th}$ face. Note that $\mathbf{F}$ is the same for all blendshape meshes. Before building the Deviation image, we need to define a few intermediate images that are useful for computations.

\begin{figure}
\begin{center}
	\includegraphics[width=0.9\linewidth]{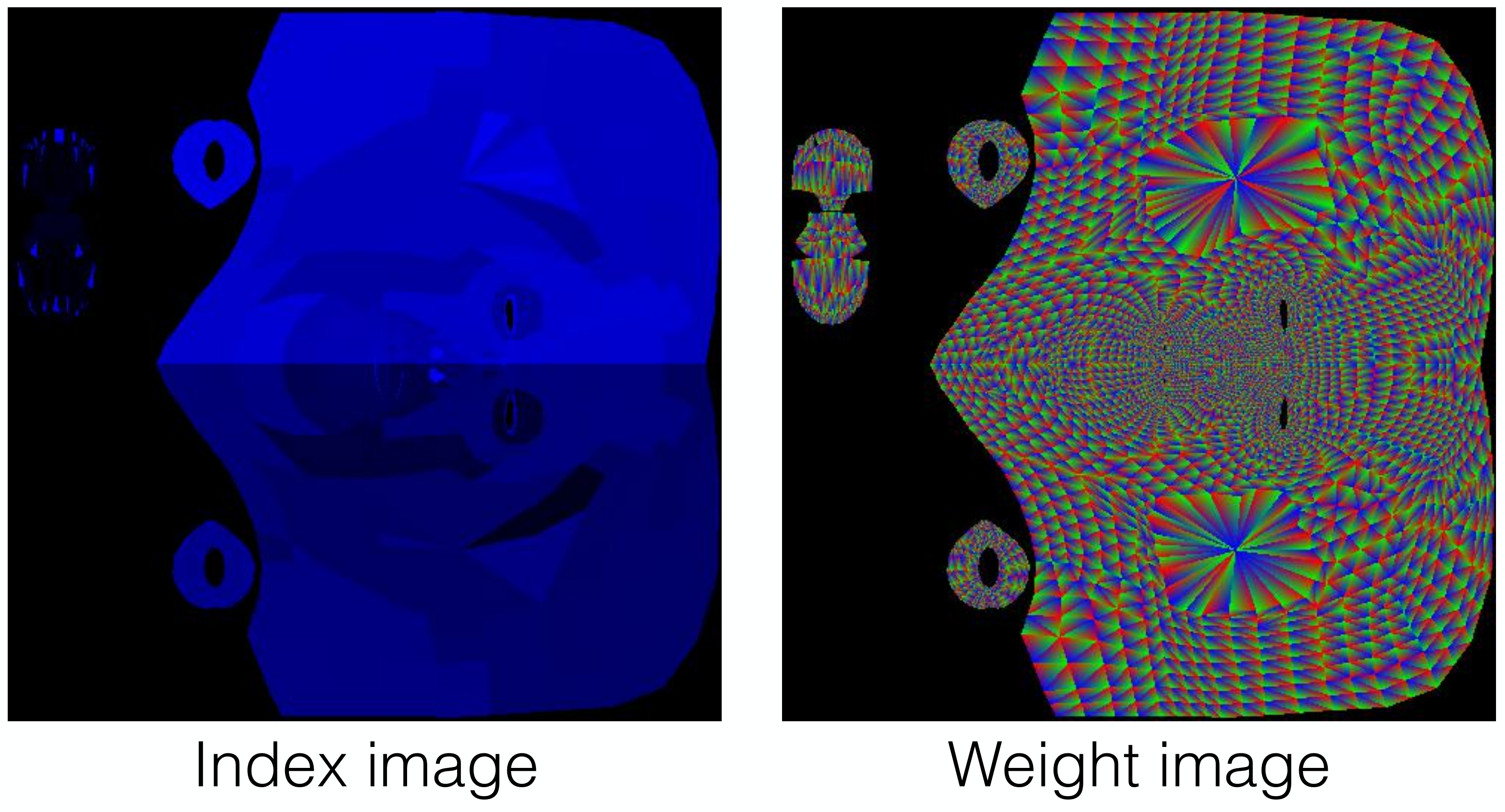}
\end{center}
   \caption{The index and weight images computed for the blendshape mesh shown in \fref{fig:template}.}
\label{fig:IndexImage}
\vspace{-0.4cm}
\end{figure}

We define an index image $\mathbf{Indx}$ such that for each pixel $(u,v)$ in the texture coordinate space, $\mathbf{Indx}(u,v)$ is the index in $\mathbf{F}$ of the triangle the pixel belongs to. We build the index image by drawing each triangle in $\mathbf{F}$ in the texture coordinate space with its own index as color. We also define a three channel weight image $\mathbf{W}$ such that $\mathbf{W}(u,v)$ is the barycentric coordinates of pixel $(u,v)$ for the triangle $\mathbf{F}(\mathbf{Indx}(u,v))$. Note that the two images $\mathbf{Indx}$ and $\mathbf{W}$ depend only on the triangulation $\mathbf{F}$ and the texture coordinates of the vertices of the blendshape meshes. They are totally independent from the user and can thus be computed once and for all and saved in the hard drive (these images are shown in \fref{fig:IndexImage}).

For each blendshape mesh, we define a vertex image $\mathbf{V}_i$ and a normal image $\mathbf{N}_i$, $i \in [0:n]$:
\begin{eqnarray*}
\begin{array}{l}
\mathbf{V}_i(u,v) = \sum\limits_{k=0}^2 \mathbf{W}(u,v)[k]\mathbf{B}_i(\mathbf{F}(\mathbf{Indx}(u,v))[k]), \\ \\
\mathbf{N}_i(u,v) = \sum\limits_{k=0}^2 \mathbf{W}(u,v)[k]\mathbf{Nmle}_i(\mathbf{F}(\mathbf{Indx}(u,v))[k]).
\end{array}
\end{eqnarray*}
We also define the difference images $\mathbf{\hat{V}}_i = \mathbf{V}_i - \mathbf{V}_0$ and $\mathbf{\hat{N}}_i = \mathbf{N}_i - \mathbf{N}_0$ for $i \in [1:n]$.

We now define our proposed Deviation image $\mathbf{Dev}$ that represents the geometric details of the user's head. Given a facial expression $x$ (\ie. $n$ blendshape coefficients), we define a blended vertex image $\mathbf{V}^x$ and a blended normal image $\mathbf{N}^x$ for the blended mesh\footnote{Note that $\mathbf{N}^x$ is not normalised. It is not a standard normal image.}:
\begin{eqnarray*}
\begin{array}{l}
\mathbf{V}^x(u,v) = \mathbf{V}_0(u,v) + \sum\limits_{i = 1}^{n}x_i\mathbf{\hat{V}}_i(u,v), \\ \\
\mathbf{N}^x(u,v) = \mathbf{N}_0(u,v) + \sum\limits_{i = 1}^{n}x_i\mathbf{\hat{N}}_i(u,v).
\end{array}
\end{eqnarray*}
Each pixel $(u,v)$ in $\mathbf{Dev}$ corresponds to the 3D point  
\begin{eqnarray}
\begin{array}{l}
\mathbf{P}^x(u,v) = \mathbf{V}^x(u,v) + \mathbf{Dev}(u,v)\mathbf{N}^x(u,v).
\end{array}
\label{eq:Deviation}
\end{eqnarray}
All values in the Deviation image are initialised to $0$\footnote{Note that $\mathbf{P}^x(u,v)$ is a linear combination of $x$.}.

Each pixel in the Deviation image represents one 3D point at the surface of the head. This drastically increases the resolution of the 3D model compared to the blendshape meshes, which is the first advantage of our proposed 3D representation. The second advantage is that details can be captured (even far from the blendshape meshes, like the hair for example). The third advantage is that a single Deviation image is sufficient to obtain detailed 3D models in all base expressions. This is because the Deviation image represents the shape parameters of the user's head (\ie, the geometric deviations to the template mesh). Moreover, as we will see in \sref{sec:Fusion}, updating the Deviation image is fast and easy. 

\begin{figure*}[t]
\begin{center}
	\includegraphics[width=1.0\linewidth]{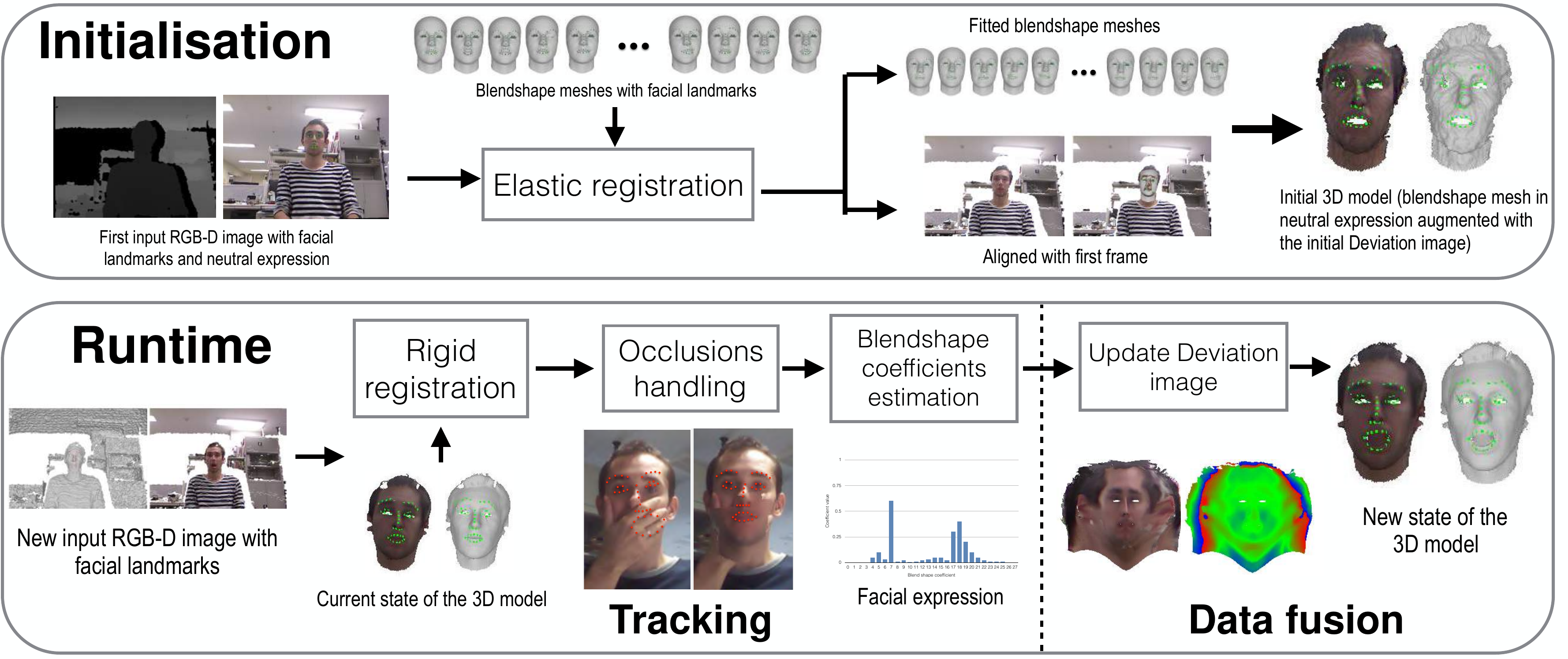}
\end{center}
   \caption{The pipeline of our proposed method. The 3D model is initialised with the first input RGB-D image that is assumed to be in neutral expression. At runtime, the rigid motion of the head is tracked and blendshape coefficients are estimated to obtain the non-rigid alignment. Then, new measurements are fused into the Deviation and color images to improve the quality of the 3D model.}
     \rule{\textwidth}{0.4pt}
\label{fig:pipeline}
\vspace{-0.4cm}
\end{figure*}

\section{Proposed 3D modeling method}

We propose a method to build a high-fidelity 3D model of the head with facial animations from a live stream of RGB-D images using our introduced 3D representation of the head. The 3D model of the head is initialised with the first input RGB-D image where
the user is assumed to be facing the camera (so that all facial features are visible) with a neutral expression. Note that this is the only constraint of our proposed method. At runtime, our built 3D model of the head is rigidly aligned to the current RGB-D image using the Iterative Closest Point (ICP) algorithm.
The blendshape coefficients are then estimated 
and the pair of Deviation and color images is updated with the current RGB-D image. The pipeline of our proposed method is illustrated in \fref{fig:pipeline}. 


\subsection{Initialisation}
We assume that the user starts facing the camera with a neutral expression (this is the only assumption done in this paper), and we initialise our 3D model with the first RGB-D image. This procedure is illustrated in the upper part of \fref{fig:pipeline} and detailed below. 

First, we detect facial features using the system called IntraFace \cite{Torre:2015}. These sparse features are matched to manually defined features in the blendshape mesh $\mathbf{B}_0$ with neutral expression. $\mathbf{B}_0$ is then scaled so that the euclidean distances between the facial features in $\mathbf{B}_0$ match the ones computed from the RGB-D image. The blendshape mesh $\mathbf{B}_0$ is aligned to the first input RGB-D image by minimizing the sum of squared distances between the matched facial features.

Second, we perform elastic registration with the facial features as proposed in \cite{Zhou:2013} to quickly and roughly fit $\mathbf{B}_0$ to the user's head. All deformations are then transferred to all other blendshape meshes $\mathbf{B}_i$, $i > 0$ \cite{Sumner:2004}. We create the Deviation and color images with the first RGB-D image (see \sref{sec:Fusion}). In order to improve tracking performances at runtime we automatically define sparse facial features in the Deviation image. We identify these features as the pixels in the Deviation image that represent the 3D points closest to the facial features detected in the first RGB-D image. Note that the quality of the initial fitting has little influence on the result because the Deviation image will compensate the initial fitting error. The most important point is that the facial features match well for accurate facial expression tracking.

\subsection{Tracking}
At runtime, we successively track the rigid motion of the head and the blendshape coefficients using the current state of our proposed 3D model of the head and the input RGB-D image. Sparse facial features are also used to improve tracking performances. This procedure is illustrated in the lower part of \fref{fig:pipeline}.

\subsubsection{Rigid head motion estimation}
\label{sec::ICP}
We estimate the pose $\begin{pmatrix} \mathbf{R} & \mathbf{t} \\ 0 &  1\end{pmatrix},$ (where $\mathbf{R}$ is the ($3 \times 3$) rotation matrix and $\mathbf{t}$ is the translation vector) of the head by computing the rigid transformation between the input RGB-D image and the 3D model of the head (that is being built) in its current expression state. We solve for this rigid alignment problem using the iterative closest point (ICP) algorithm \cite{Rusinkiewicz:2001}, which is based on dense point-to-plane constraints on the depth image\footnote{We decide not to use the facial features for the rigid tracking because they can be unreliable when part of the face is occluded.}. 
We eliminate correspondences that are farther than $1$ cm and those that normal vectors have difference in angle greater than $30$ degrees. For each pose estimation, we run $6$ iterations of the ICP.

We optimise the transformation parameters using the Lie algebra {\bf se}(3) to represent 3D transformations of {\bf SE}(3) \cite{Lie} (this gives a robust solution). A 3D transformation matrix $\mathbf{T}$ (of dimension $4 \times 4$) is represented by a 6-vector $\mathbf{l}$ in {\bf se}(3), which is mapped to the matrix $\mathbf{T(l)}$ using the exponential function.

Our point-to-plane fitting term on the depth image is:
\begin{eqnarray*}
\begin{array}{l}
r_{(u,v)}^S(\mathbf{l}) = \alpha(u,v)(\mathbf{n}_{(u,v)}(\mathbf{T(l)}\mathbf{P}^x(u,v) - \mathbf{v}_{(u,v)}))^2,
\end{array}
\label{eq::pperr}
\end{eqnarray*}
where $(u,v)$ is a pixel in the Deviation image, $\mathbf{v}_{(u,v)}$ is the closest point to $\mathbf{T(l)}\mathbf{P}^x(u,v)$\footnote{Note that by abuse of notations $\mathbf{P}^x(u,v)$ is here considered as the 4-vector homogeneous coordinates of $\mathbf{P}^x(u,v)$ defined in \eqref{eq:Deviation}} in the depth image, $\mathbf{n}_{(u,v)}$ is the normal vector of $\mathbf{v}_{(u,v)}$ and $\alpha(u,v)$ is a binary visibility term defined in \sref{seq::occlusions}. Note that $\mathbf{v}_{(u,v)}$ is obtained by projecting $\mathbf{T(l)}\mathbf{P}^x(u,v)$ in the depth image. This allows fast computations on the GPU.

The pose of the head $\mathbf{T(l^*)}$ is computed by solving the minimisation problem for the total fitting term:
\begin{eqnarray}
\begin{array}{l}
l^* = \arg\min\limits_{\mathbf{l}} \sum\limits_{(u,v)}r_{(u,v)}^S(\mathbf{l}). 
\end{array}
\label{eq::minICP}
\end{eqnarray}
We solve \eqref{eq::minICP} using the Levenberg-Marquardt algorithm. 

\begin{figure*}[t]
\begin{center}
	\subfloat[The candidate pixels are those that belong to the 2D projected segment (red segment) in the input RGB-D image.] {\includegraphics[width=0.47\linewidth]{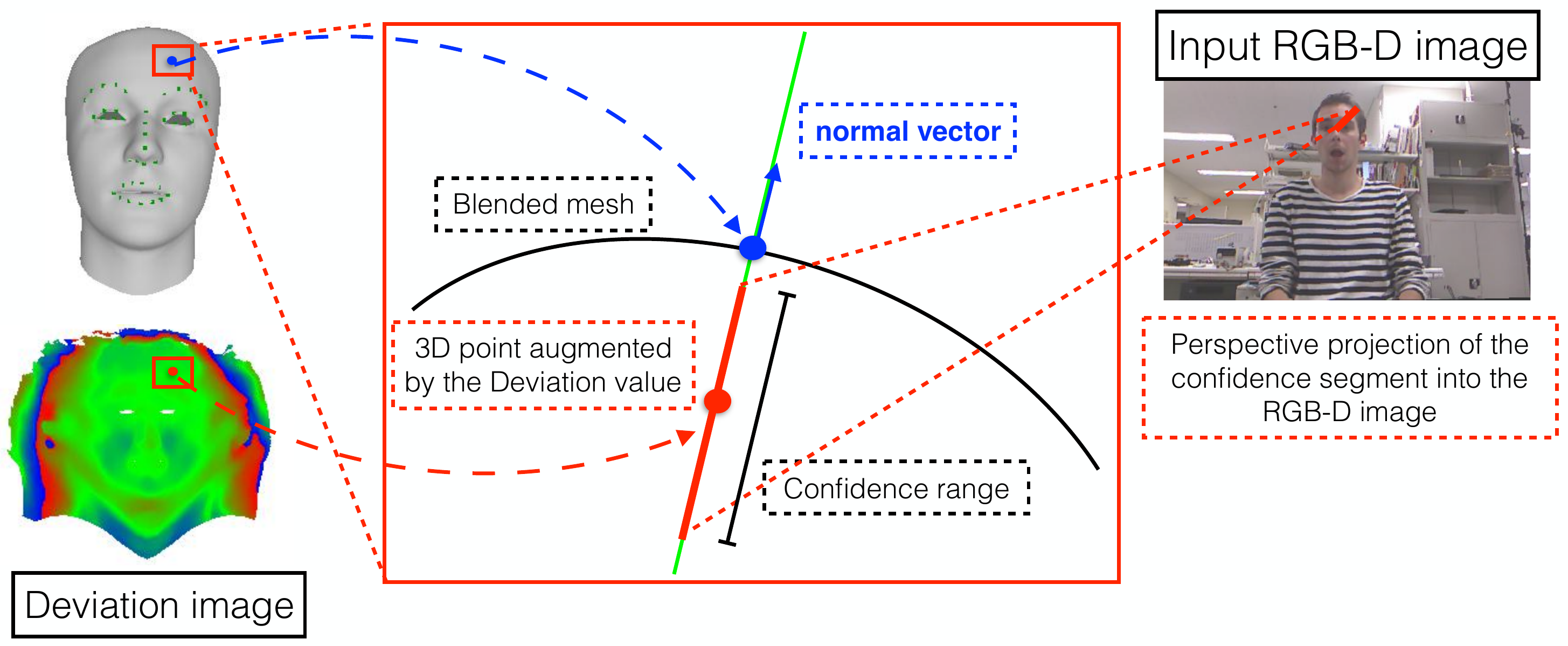}}
\qquad	
	\subfloat[The point closest to the normal vector is identified as the best match.] {\includegraphics[width=0.47\linewidth]{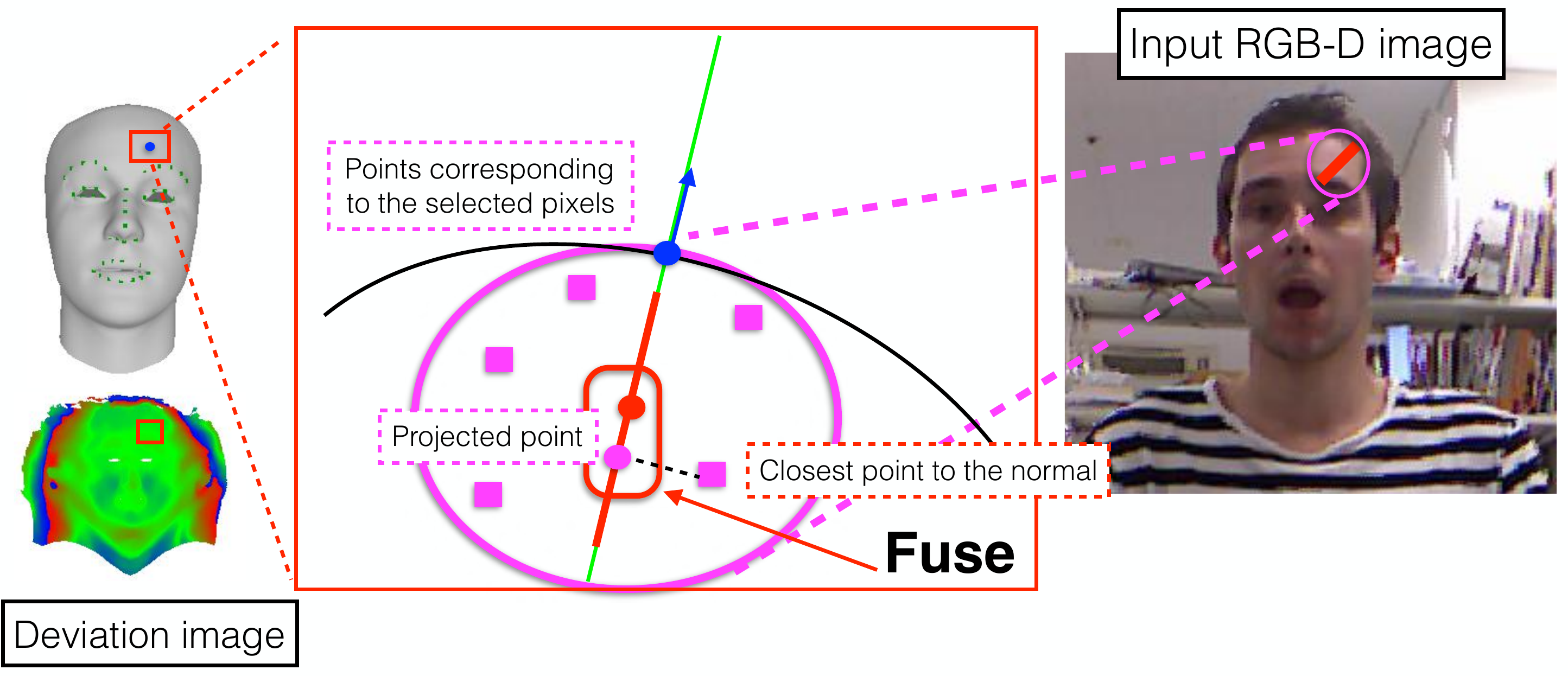}}
\end{center}
\caption{Data fusion into the Deviation image. For each pixel in the Deviation image, (a) a few candidate points are selected in the input image and (b) the best match among these candidates is used to update the pixel value.}
     \rule{\textwidth}{0.4pt}
\label{fig:fusionb}
\vspace{-0.4cm}
\end{figure*}

\subsubsection{Blendshape coefficients estimation}
For each input RGB-D image we estimate the blendshape coefficients using the same approach as in \cite{Hsieh:2015}. Note that differently from \cite{Hsieh:2015} we used the point-to-point constraints on the 3D facial features (instead of on the 2D facial features). Moreover, we use all points available from the Deviation image for dense point correspondences. Our point-to-plane fitting term on the depth image is
\begin{eqnarray*}
\begin{array}{l}
c_{(u,v)}^S(x) = (\mathbf{n}_{(u,v)}(\mathbf{T(l^*)}\mathbf{P}^x(u,v) - \mathbf{v}_{(u,v)}))^2,
\end{array}
\end{eqnarray*}
where $(u,v)$ is a pixel in the Deviation image, $\mathbf{v}_{(u,v)}$ is the closest point to $\mathbf{T(l^*)}\mathbf{P}^x(u,v)$ in the depth image and $\mathbf{n}_{(u,v)}$ is the normal vector of $\mathbf{v}_{(u,v)}$.

Our point-to-point fitting term on 3D facial features is 
\begin{eqnarray*}
\begin{array}{l}
c_{j}^F(x) = \|\mathbf{T(l^*)}\mathbf{P^x}(lmk_j) - \mathbf{v}_{j}\|_2^2,
\end{array}
\end{eqnarray*}
where $lmk_j$ is the location of the $j^{th}$ landmark in the Deviation image and $v_{j}$ is the $j^th$ 3D facial landmark in the RGB-D image.

The blendshape coefficients are computed by solving the minimisation problem for the total fitting term
\begin{eqnarray}
\begin{array}{l}
x^* = \arg\min\limits_x \sum\limits_{(u,v)}c_{(u,v)}^S(x) + w_L\sum\limits_{j} c_{j}^F(x) + \\ w_S(\sum\limits_{k=1}^{n}x_k^2  + \sum\limits_{k=1}^{n}(x_k- \hat{x_k})^2),
\end{array}
\label{eq::blendcoeff}
\end{eqnarray}
where $w_L$ and  $w_S$ are two weighting factors set to $100$ and $0.0004$ in our experiments, and $\{\hat{x_k}\}_{k \in[1:n]}$ are the previous blendshape coefficients.

We solve eq. \eqref{eq::blendcoeff} using an iterative projection method. At each iteration eq. \eqref{eq::blendcoeff} is linearized around the current estimate of the blendshape coefficients. The updated coefficients are computed using the parallel relaxation algorithm 2.1 of \cite{Sugimoto:1995}. The updated coefficients are clamped to the interval $[0,1]$ before proceeding to the next iteration. In our experiments, we used $6$ iterations.

\subsection{Data fusion}
\label{sec:Fusion}
Our proposed 3D model of the head grows and refines online with input RGB-D images. We use a running median strategy to integrate new RGB-D data into the Deviation and color images. To do so, we prepare for each pixel in the Deviation image a sorted list of deviation values, initialized with the empty list and with maximum size $100$. At any time the value of a pixel in the Deviation image is taken as the current median of the corresponding list. At run time, for each pixel in the Deviation image, the corresponding point in the input RGB-D image is selected and a deviation value is inserted in place in the sorted list (if there is a valid correspondence). If the list reaches the maximum size, then the element farthest from the median is removed. How to select the corresponding point is explained below. Note that the color value is obtained by projecting the 3D point (in $\mathbf{P}^x$) into the input RGB image. 

For a given $x$ (blendshape coefficients), each pixel $(u,v)$ in the Deviation image corresponds to a 3D point $\mathbf{P}^x(u,v)$ that lies in the line $L^x(u,v)$ directed by the vector $\mathbf{N}^x(u,v)$ and passing by the 3D point $\mathbf{V}^x(u,v)$ (see \eref{eq:Deviation}). For each input RGB-D image, with estimated pose $\mathbf{T(l^*)}$ (with rotation $\mathbf{R}$) and blendshape coefficients $x$, we update the list of deviation values for all pixels using the 3D point in the RGB-D image that is closest to the line $\hat{L^x}(u,v)$, directed by the vector $\mathbf{R}\mathbf{N}^x(u,v)$ and passing by the 3D point $\mathbf{T(l^*)}\mathbf{V}^x(u,v)$. This procedure is illustrated in \fref{fig:fusionb} and detailed below\footnote{Note that  $\mathbf{Dev}$ records deviation in the normal direction. This is why we must select data in the normal direction for consistency.}. 

For each pixel $(u,v)$, we search for the 3D point in the RGB-D image that is closest to the line $\hat{L^x}(u,v)$ by walking through a projected segment in the depth image. We define the segment $S(u,v) = [\mathbf{T(l^*)}\mathbf{P}^x(u,v) - \lambda \mathbf{R}\mathbf{N}^x(u,v); \mathbf{T(l^*)}\mathbf{P}^x(u,v) + \lambda \mathbf{R}\mathbf{N}^x(u,v)]$, where $\lambda = 5$ cm if the list of deviation values is empty (in such a case $\mathbf{Dev}(u,v) = 0$), $\lambda = \max(1, \frac{5}{s})$ cm otherwise (where $s$ is the current size of the list). We then walk through the projected segment $\pi(S(u,v))$, where $\pi$ is the perspective projection operator and identify the point $\mathbf{p}_{u,v}$ closest to the line $\hat{L^x}(u,v)$. We compute the distance $d_{(u,v)}$ from $\mathbf{p}_{u,v}$ to the corresponding point $\mathbf{T(l^*)}\mathbf{V}^x(u,v)$ on the blended mesh in the direction $\mathbf{R}\mathbf{N}^x(u,v)$: 
\begin{eqnarray*}
\begin{array}{l}
d_{(u,v)}= (\mathbf{p}_{u,v} - (\mathbf{T(l^*)}\mathbf{V}^x(u,v))) \cdot (\mathbf{R}\mathbf{N}^x(u,v)),
\end{array}
\end{eqnarray*}
where $\cdot$ is the scalar product. $d_{(u,v)}$ is then inserted in place in the sorted list of deviation values that corresponds to pixel $(u,v)$.

We do not update the value of the Deviation image at pixel (u,v) when the corresponding point $\mathbf{p}_{u,v}$ is either farther than $1$ cm to the line $\hat{L^x}(u,v)$, farther than $\tau$ cm to the point $\mathbf{P}^x(u,v)$ (with $\tau = 3$ if the list is empty and $\tau = 1$ otherwise), or when the difference in angle between the normal vector of $\mathbf{p}_{u,v}$ and $\mathbf{N}^x(u,v)$ is greater than $45$ degrees. 

At each frame, we apply a bilateral gaussian filter (with a window size of $3 \times 3$ pixels) to the Deviation image to remove outliers. 

\section{Occlusions}
\label{seq::occlusions} 
We detect occlusions in the input RGB-D images following the strategy proposed in \cite{Hsieh:2015}. Both depth and color information are combined to identify occluded pixels and vote in a superpixel space. The obtained segmented RGB-D image is completed using our built 3D model of the head to improve facial features detection and blendshape coefficients estimation.    
 
%

\subsection{Face segmentation}
\label{seq::segmentation} 

 \begin{figure*}[t]
\begin{center}
	\includegraphics[width=1.0\linewidth]{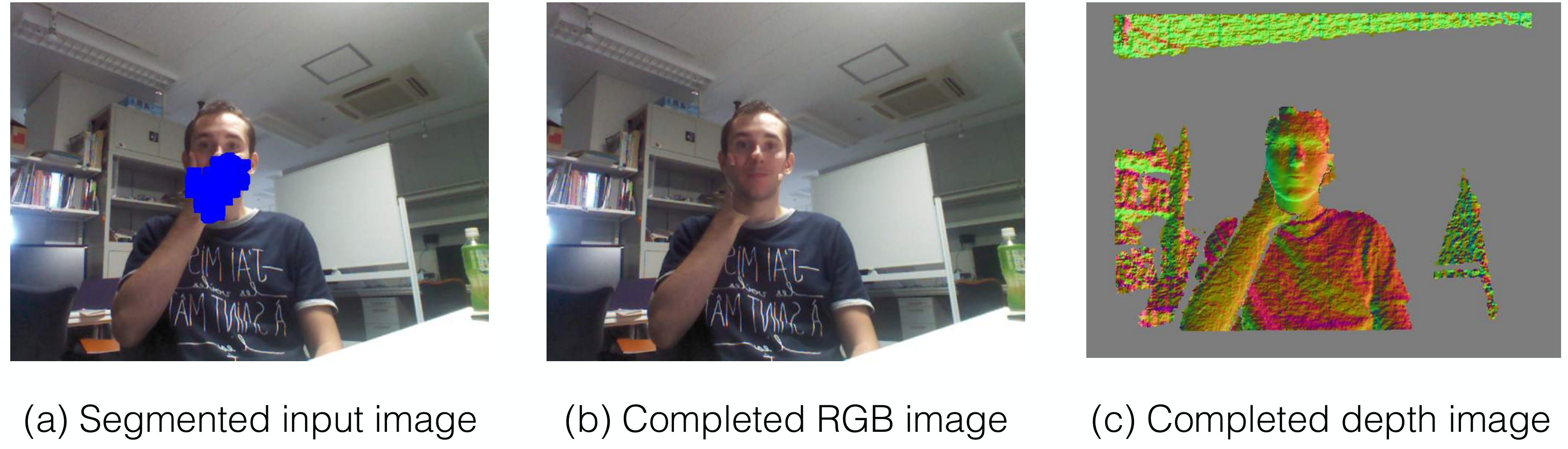}
\end{center}
   \caption{We complete the occluded regions of the input RGB-D image in both RGB and depth channels using our built 3D model of the head. The completed depth image is used to stably and robustly track facial expressions even when a large part of the face remains occluded for a long period of time.}
     \rule{\textwidth}{0.4pt}
\label{fig::seg2}
\vspace{-0.4cm}
\end{figure*}


We segment the face in the input RGB-D image into occluded and non-occluded regions using the 3D model of the head in its current pose and expression state. 
To each pixel $(u,v)$ in the Deviation image is associated a binary value $\alpha (u,v)$ (used in \sref{sec::ICP}) that determines if the corresponding point in the Deviation image is occluded or not. 


For each input image, the region of the face is segmented into Superpixels \cite{SLIC}.
For each Superpixel we count the number of inliers and outliers, identified using the difference in depth values and color values between the rendered 3D model of the head and the input RGB-D image. Namely, a pixel $(u,v)$ is considered as an inlier if (
the captured data is not in front of the built 3D model of the head (with a threshold $\tau_d$ set to $1$ cm in our experiments), and if the difference of colors in the CIELAB colorspace between the input and the rendered color images is less than a threshold $\tau_c$ set to $40$ in our experiments. Thereafter, each superpixel is considered as occluded if the number of outliers is greater than the number of inliers. 
An example of a segmentation result is shown in \fref{fig::seg2} (a).

For each input RGB-D image, after face segmentation is done, each pixel in the Deviation image is projected into the input RGB-D image, and its occlusion value $\alpha(u,v)$ is set to $0$ if the corresponding pixel is occluded, $1$ otherwise.

\subsection{Occlusion completion}
Parts of the face in the input image that are occluded should not be used for head pose tracking and expression capture. For the head pose tracking, as explained in \sref{sec::ICP} (eq. \eqref{eq::pperr}) we use only non-occluded pixels in the energy function. Unfortunately, using the same simple strategy to capture the facial expression does not work. This is because:
\begin{itemize}
\item The detected facial features can become totally wrong because of occlusions, which will corrupt the energy function in eq. \eqref{eq::blendcoeff}. Note that even the non-occluded facial features may become corrupted when part of the face is occluded.
\item Several blendshape coefficients do not have any influence anymore when the part they control is occluded. This may lead to unstable and unrealistic results.
\end{itemize}

Replacing the occluded regions with the corresponding parts in the rendered image, and using this completed image to re-detect facial features is an efficient way to overcome the first problem discussed above. By contrast with \cite{Hsieh:2015} where only the RGB image is completed, we also complete the depth image using our detailed 3D model of the head.
After completion, we obtain a completed RGB-D image (see \fref{fig::seg2}) that is used to re-detect the facial features and to optimise the blendshape coefficients in eq. \eqref{eq::blendcoeff}. Note that small errors in the estimated expressions would be propagated into the next completed image. This would result into drifting errors and unpleasant results after a few frames. Therefore we update the Deviation image, the blended vertex image ($\mathbf{V}^x$) and the blended normal images ($\mathbf{N}^x$) only for non-occluded pixels. 
By doing so we can obtain stable facial expression tracking even when large parts of the face remain occluded for a long time. 

\begin{figure*}[t]
\begin{center}
	\includegraphics[width=1.0\linewidth]{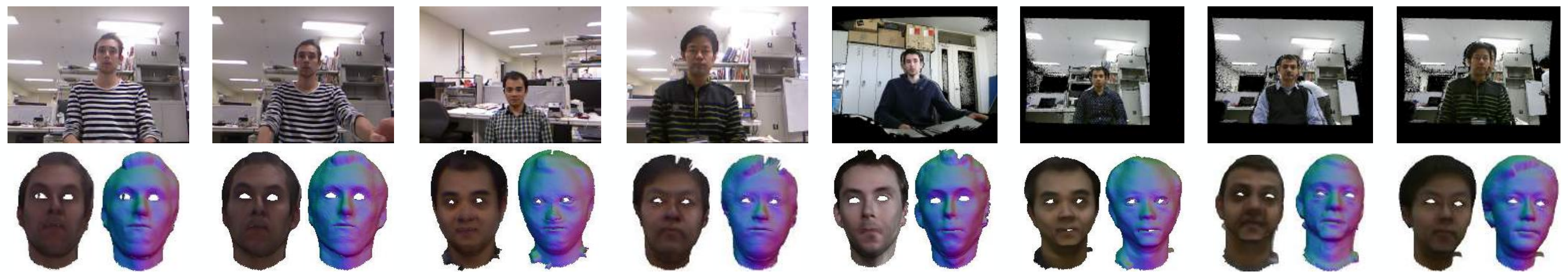}
\end{center}
   \caption{The first frame and final retargeted 3D model from results obtained with our proposed method shown in the supplementary video. The four results on the left side of the figure were obtained using a Kinect for XBOX 360, while the four results on the right side of the figure were obtained with a Kinect V2.}
     \rule{\textwidth}{0.4pt}
\label{fig:res1}
\vspace{-0.4cm}
\end{figure*}

\section{Results}

\begin{figure*}
\begin{center}
\qquad	
	\subfloat {\includegraphics[width=0.9\linewidth]{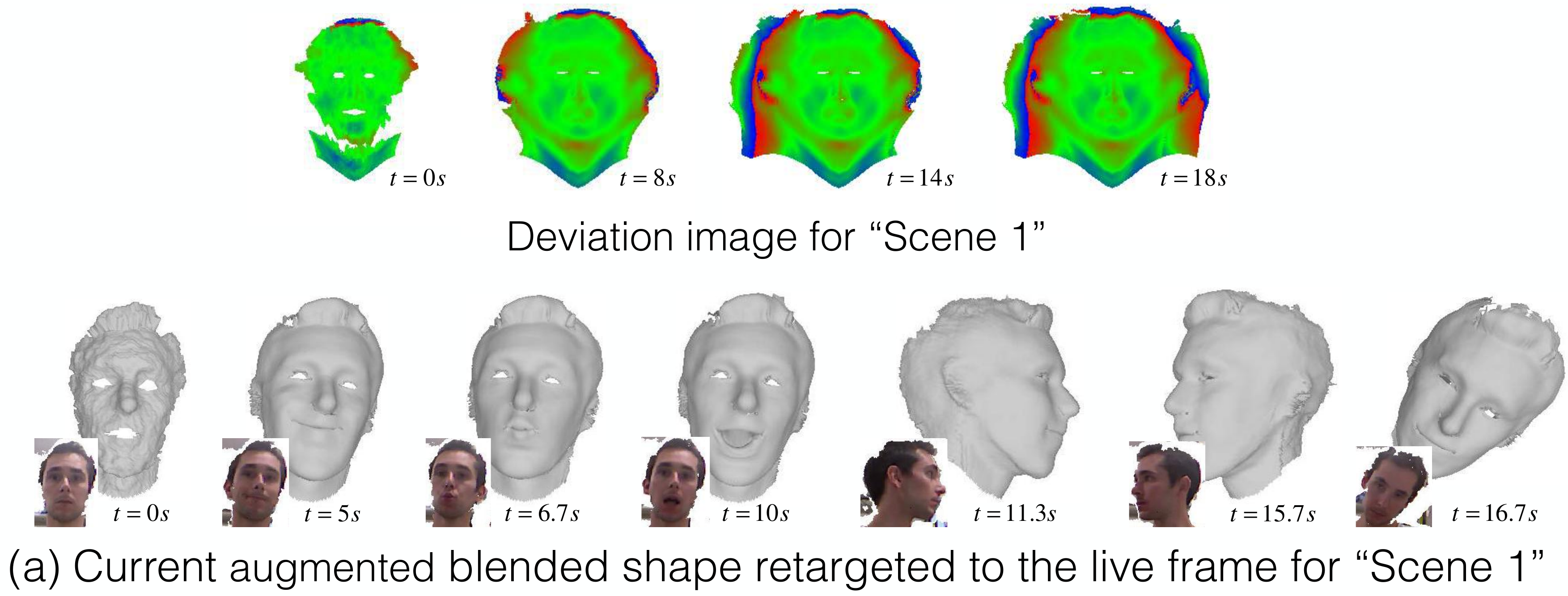}}
\qquad	
	\subfloat {\includegraphics[width=0.9\linewidth]{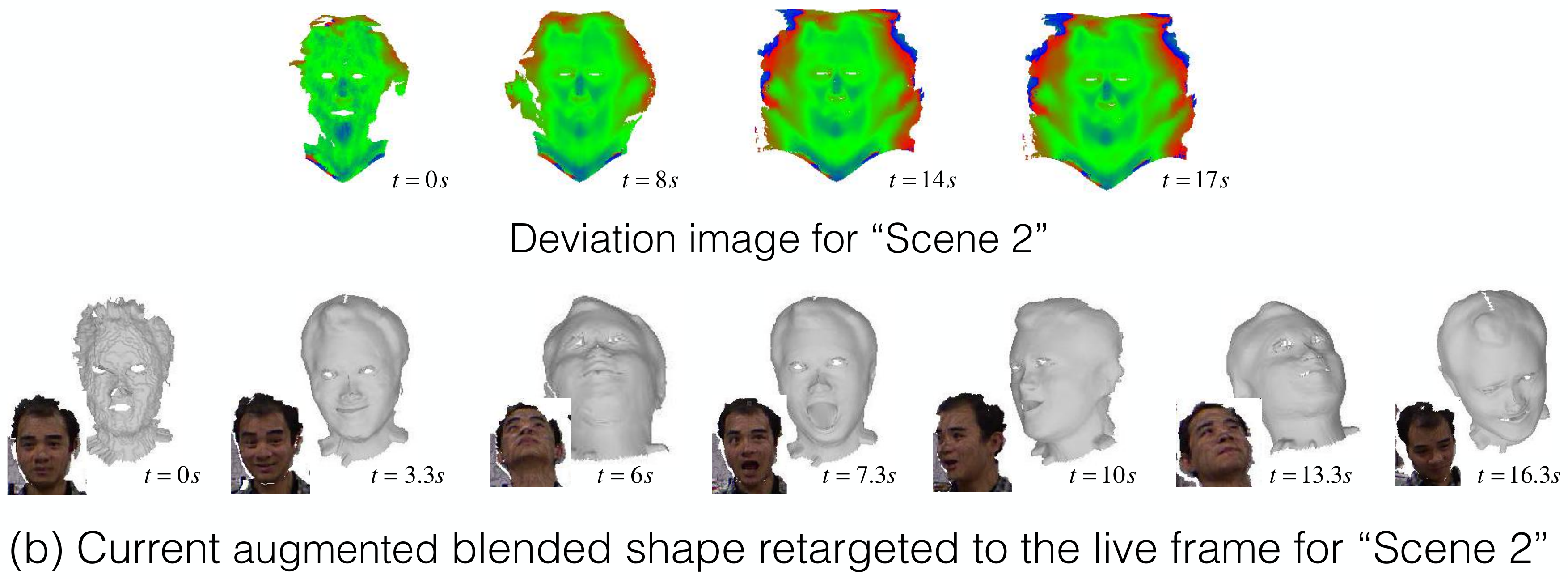}}
\qquad	
	\subfloat {\includegraphics[width=0.9\linewidth]{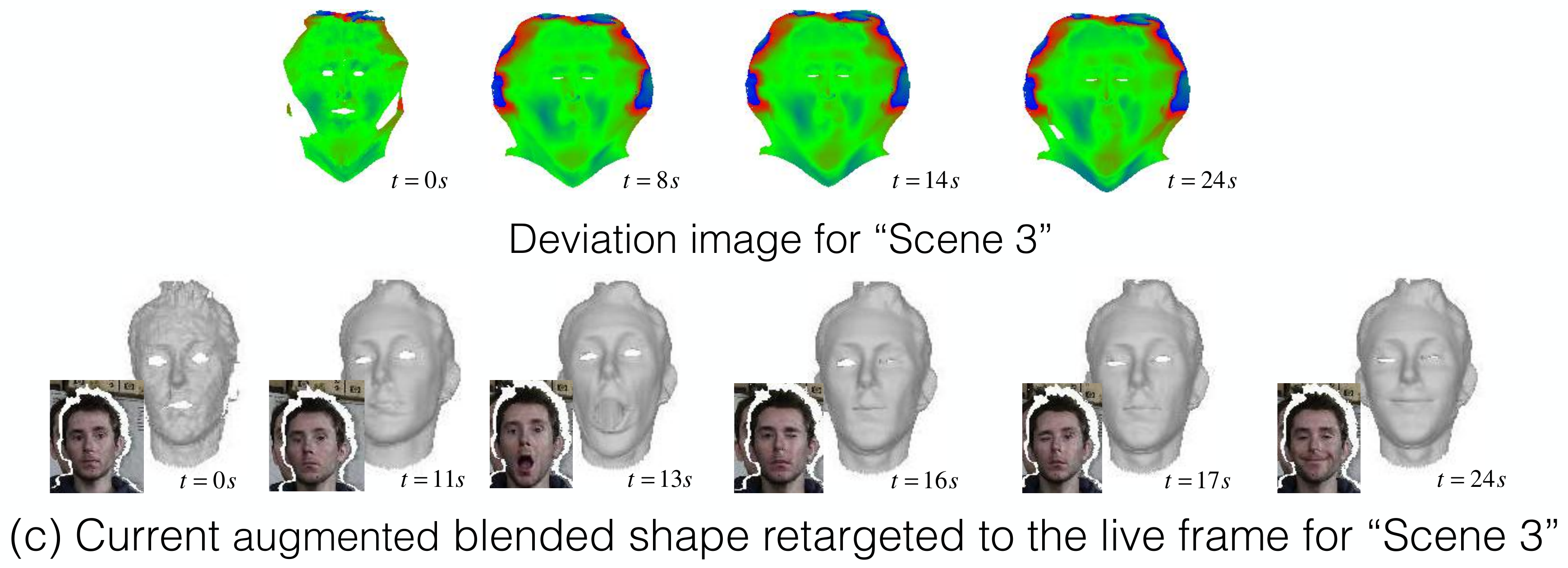}}
\end{center}
\caption{Results obtained with our proposed method for three scenes captured with a Kinect V1 ((a) and (b)) and with a Kinect V2 (c). Upper rows of (a), (b) and (c) show the Deviation images as they grow and refine over time. Lower rows show the augmented blended meshes at different time (with the current Deviation image).}
\label{fig:res2}
\end{figure*}

We demonstrate the ability of our proposed method to generate high-fidelity 3D models of the head in dynamic scenes along with the facial motions, with real experiments using both the Kinect V1 and Kinect V2 sensors. The Kinect V1 sensor (based on structured light) provides RGB-D images at $30$ fps with a resolution of $640 \times 480$ pixels; the Kinect V2 sensor (based on time of flight) provides RGB-D images at $30$ fps with a resolution of $1920 \times 1080$ pixels in the color image and of $512 \times 424$ pixels in the depth image. 

In all our experiments, we used a Deviation and color image with resolution of $240 \times 240$ pixels (\ie, with average distance between neighbouring points of about $1$ mm). Our proposed method has four parameters: $w_L$, $w_S$, $\tau_d$ and  $\tau_c$\footnote{Note that we used standard parameters for the ICP algorithm.}. We set these parameters experimentally (\ie, $w_L=100$, $w_S=0.004$, $\tau_d=1$ cm and $\tau_c=40$) and fixed them through all our experiments.

The full pipeline of our proposed method runs at about $15$ fps on a macbook pro with a $2.8$ GHz Intel Core i7 CPU with $16$ GB RAM and an AMD Radeon R9 M370X graphics processor. While our code is not fully optimised, we measured the following average timings: head pose estimation took about $2$ ms, blendshape coefficients estimation took about $20$ ms, data fusion took about $7$ ms and occlusions handling (\ie, input RGB-D image segmentation, completion and facial features re-detection) took about $40$ ms. 

In all our experiments, the user is moving in front of the camera. As a consequence, static 3D reconstruction techniques such as using a laser scanner cannot be used. Therefore we do not have the ground truth to perform quantitative evaluation of the proposed method, and we only report qualitative evaluation.

\Fref{fig:res1} shows the first frame of RGB-D videos captured with both sensors, as well as the final 3D model obtained with our proposed method, in the pose of the first frame and with neutral expression. These videos illustrate several challenging situations with various facial expressions, extreme head poses and different shapes of the head. Our proposed Deviation image that augments the blendshape meshes allowed us to capture detailed and various geometric details around the head, including the hair (which was not possible with state-of-the-art blendshape methods \cite{Hsieh:2015}), with similar accuracy for different users. In addition, the (underlying) blendshape representation allowed us to capture fine facial motions in real-time, which also helped to build accurate 3D models of the head even in dynamic scenes. Our proposed method is robust to data noise, head pose and facial expression changes, which allowed us to obtain similarly satisfactory results with different sensors.

In \fref{fig:res2}, we can see that the Deviation image grows and refines over time to generate accurate 3D models with various facial expressions. In particular, by using facial features in addition to the depth image, we could successfully track the movement of the eyelids (\fref{fig:res2} (c) at $t = 17s$, $t = 22s$ and $t=24s$). This is not possible without using facial features because the depth information alone can not distinguish between "eye closed" and "eye opened" (\cite{Newcombe:2015}). Furthermore, contrary to \cite{Newcombe:2015} we do not need to start the sequence by scanning the head with mouth opened because we know (with the blendshape meshes) the topology of the head (\ie, mouth and eyes can open and close).

\begin{figure*}
\begin{center}
\qquad	
	\subfloat {\includegraphics[width=0.45\linewidth]{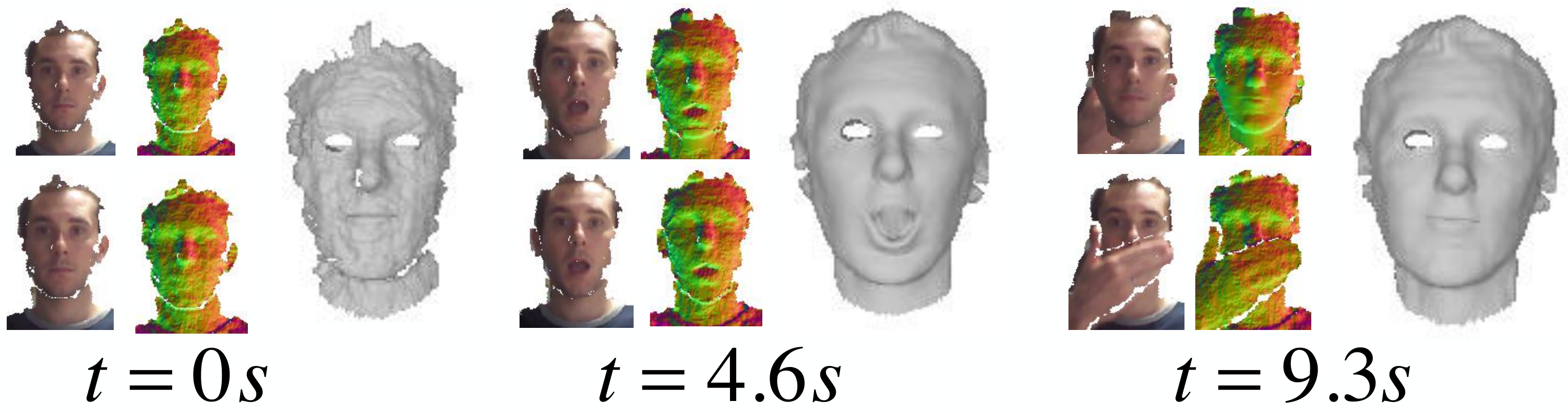}}
\qquad	
	\subfloat {\includegraphics[width=0.47\linewidth]{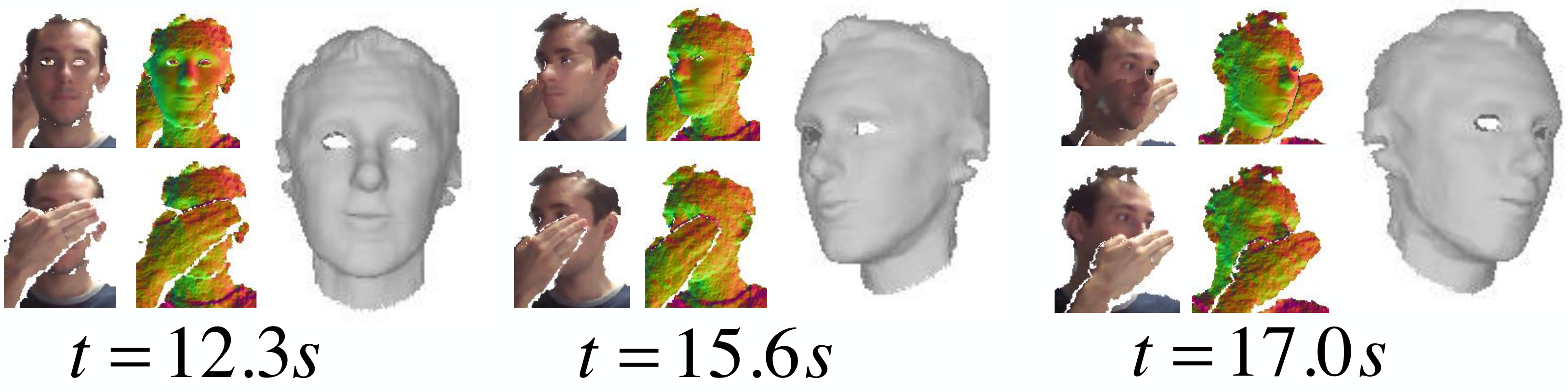}}
\end{center}
\caption{Real-time reconstruction of animated 3D models of the head with occlusions. For each result we show on the bottom left the input color and normal images, on the upper left the completed color and normal images and on the right the built 3D model of the head in its current pose and facial expression state.}
\label{fig:resocclusions}
\vspace{-0.2cm}
\end{figure*}

\begin{figure*}
\begin{center}
\qquad	
	\subfloat {\includegraphics[width=0.45\linewidth]{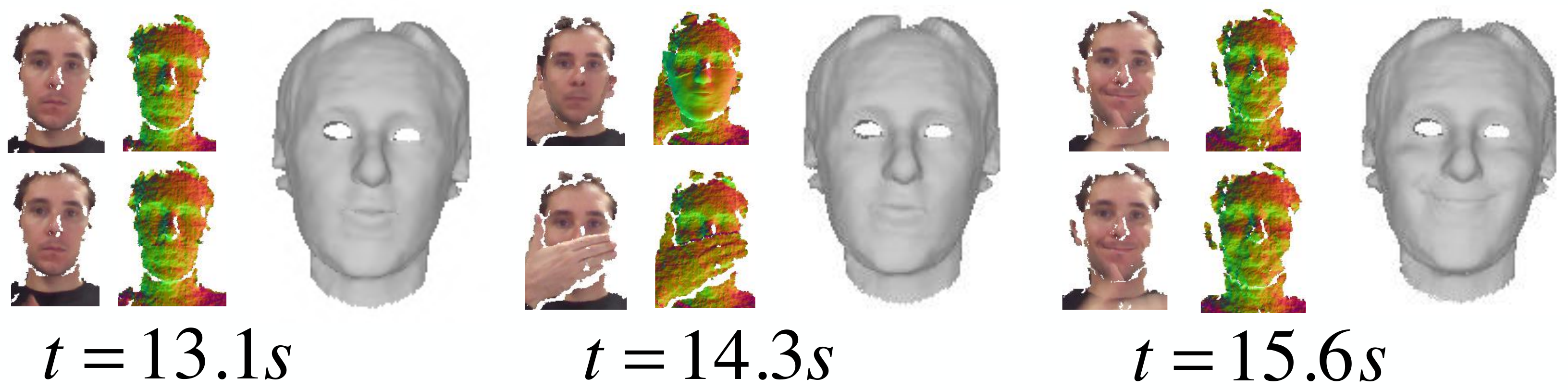}} 
\qquad	
	\subfloat {\includegraphics[width=0.45\linewidth]{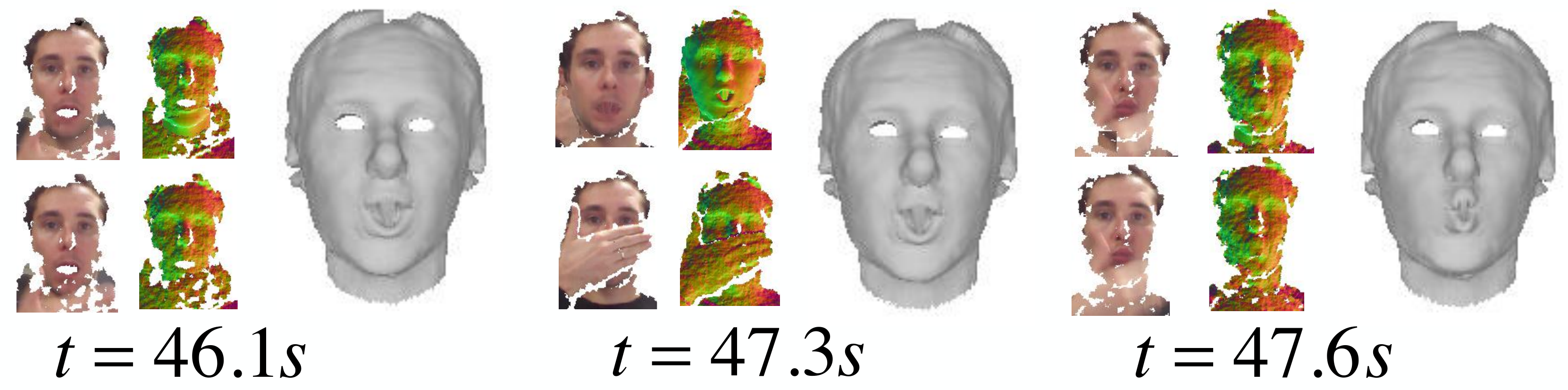}}
\end{center}
\caption{Real-time reconstruction of animated 3D models of the head with occlusions in situations where the expression of the face is changing while being occluded. For each result we show on the bottom left the input color and normal images, on the upper left the completed color and normal images and on the right the built 3D model of the head in its current pose and facial expression state.}
\label{fig:resocclusions2}
\vspace{-0.2cm}
\end{figure*}

\Fref{fig:resocclusions} shows results obtained with our proposed method in situations where a large part of the face is occluded. As we can see from these results, our proposed method could successfully reconstruct the detailed 3D model of the head, track the pose of the head and capture the facial expressions. Even when large parts of the face are occluded for a long time, we could still accurately track the pose of the head and we obtained stable and natural facial expressions. \Fref{fig:resocclusions2} shows results obtained with our proposed method in situations where the facial expression changes while the part that is changing is occluded. Even in these challenging situations, our proposed method could correctly estimate the facial expressions at the end of each occlusion sequence. 

\begin{figure}
\begin{center}
	\subfloat {\includegraphics[width=0.38\linewidth]{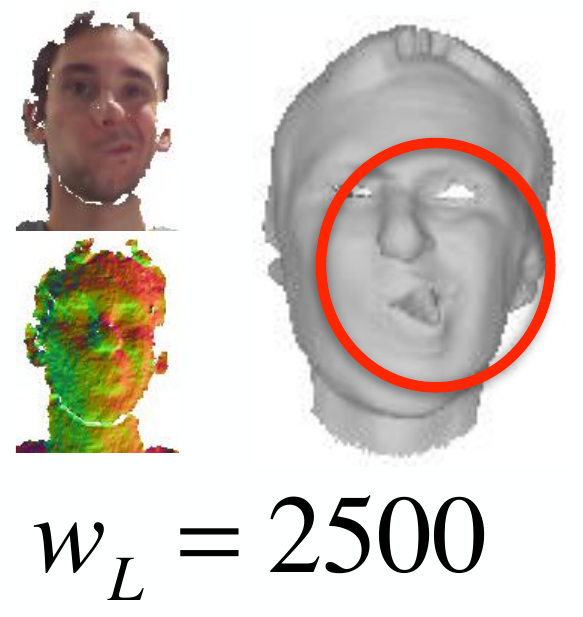}}
~~~	\subfloat {\includegraphics[width=0.45\linewidth]{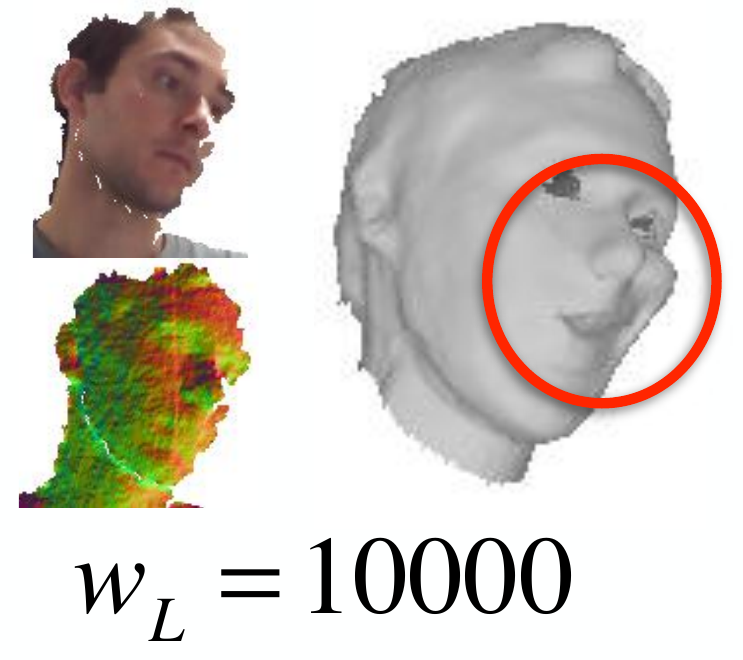}}
\end{center}
\caption{Results obtained with our proposed method when the parameter $w_L$ was set to $2500$ and $10000$. The circled areas show the parts of the face that were the most affected by the changes in the parameter's value. Setting a too high value for $w_L$ leads to unstable results when facial features are wrongly detected.}
\label{fig:resparamwL}
\vspace{-0.2cm}
\end{figure}

\begin{figure*}
\begin{center}
\qquad	
	\subfloat[A low value for $w_S$ allows to track fine motions.] {\includegraphics[width=0.45\linewidth]{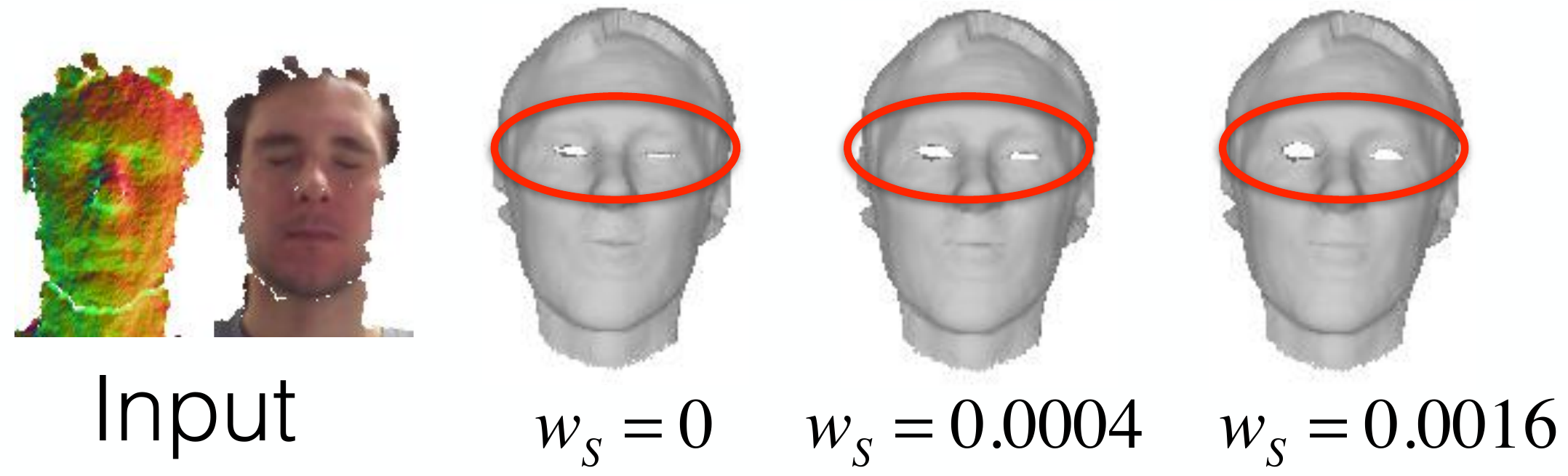}}
\qquad	
	\subfloat[A high value for $w_S$ improves robustness and stability.] {\includegraphics[width=0.45\linewidth]{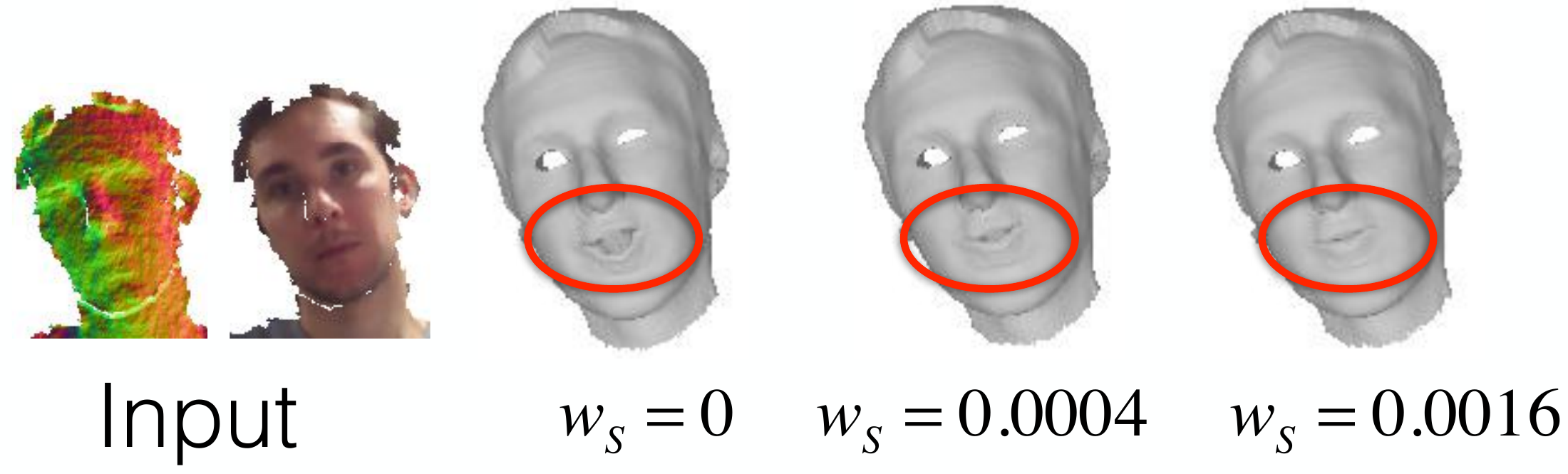}}
\end{center}
\caption{Results obtained with our proposed method when the parameter $w_S$ was set to $0.0$, $0.0004$ and $0.0016$. The circled areas show the parts of the face that were the most affected by the changes in the parameter's value.}
\label{fig:resparamw}
\vspace{-0.2cm}
\end{figure*}

\begin{figure*}
\begin{center}
\qquad	
	\subfloat[A high value for $\tau_d$ leads to missing occluded areas of the face, which leads to unsatisfactory results.] {\includegraphics[width=0.9\linewidth]{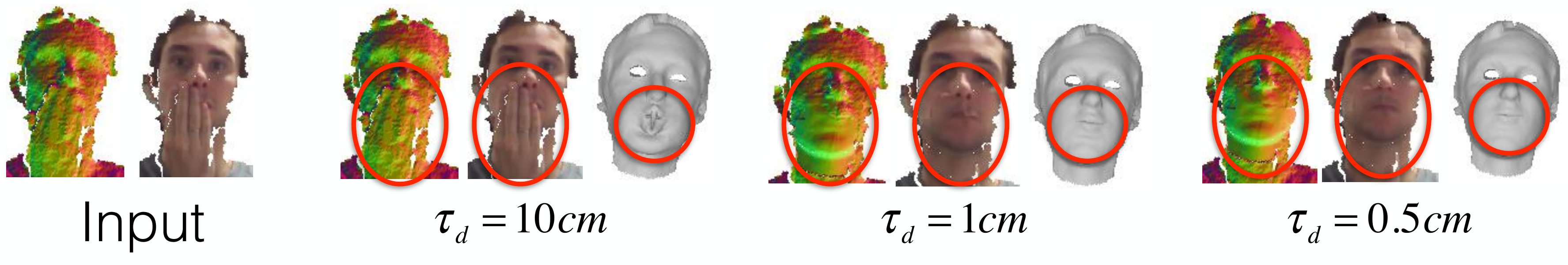}}
\qquad	
	\subfloat[A low value for $\tau_d$ leads to detecting non-occluded parts of the face as occluded. Some facial expression changes were missed.] {\includegraphics[width=0.9\linewidth]{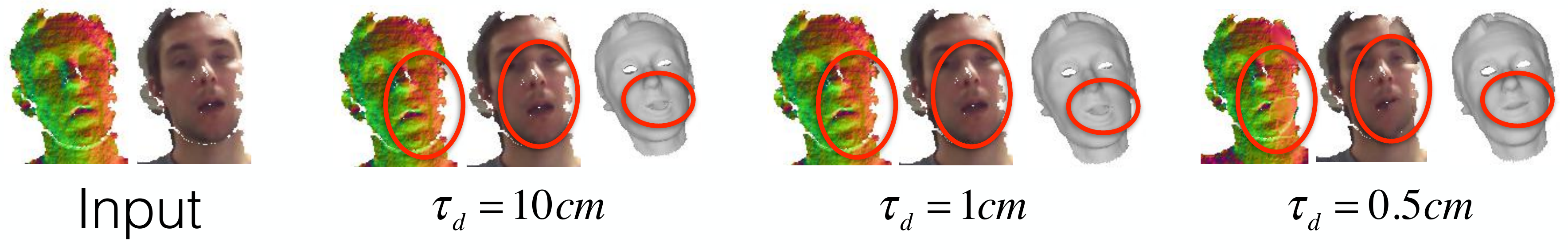}}
\end{center}
\caption{Results obtained with our proposed method when the parameter $\tau_d$ was set to $10$ cm, $1$ cm and $0.5$ cm. The circled areas show the parts of the face that were the most affected by the changes in the parameter's value.}
\label{fig:resparamtau}
\vspace{-0.2cm}
\end{figure*}

 \Fref{fig:resparamwL}, \fref{fig:resparamw}  and  \fref{fig:resparamtau} show the effects of changing the parameters of our proposed method on the results. \Fref{fig:resparamwL} illustrates the effects of varying the value of parameter $w_L$. As we can see, setting a too high value for the weight for facial landmarks in eq. \eqref{eq::blendcoeff} may results in unnatural results on frames where the facial features were wrongly estimated. On the contrary setting a too low value may prevent our proposed method to track fast motions. We found that a value of $w_L = 100$ gives a good compromise between robustness to errors in facial feature estimation and fast motion tracking. 
 
 \Fref{fig:resparamw} illustrates the impact of parameter $w_S$ on the results. As we can see from these results, setting $w_S$ to a low value ($0.0$ for example) allows to track fine motions like eyes closing (see the circled areas of \fref{fig:resparamw} (a)). The drawback, however, is that the results may become unstable, as shown in the circled areas of \fref{fig:resparamw} (b). We found that a value of $w= 0.004$ gives a good compromise between stability and accuracy for facial expression estimation. 
 
 \Fref{fig:resparamtau} shows the impact of parameter $\tau_d$ (that controls the quality of the face segmentation) on the results. As expected a high value of $\tau_d$ results in occluding parts not being detected, and thus produces unsatisfactory results when parts of the face are occluded (see \fref{fig:resparamtau} (a)). On the contrary, a too low value of $\tau_d$ results in non-occluded parts of the face being detected as occluded due to data noise. As shown in the circled area of \fref{fig:resparamtau} (b), this may lead to missing some changes in facial expression. Similar effects are observed when varying the value of parameter $\tau_c$. We found that a value of $\tau_d = 1$ cm and $\tau_c = 40$ gives good results for face segmentation and lead to the best results in our experiments. Note that finding an automatic way to set the optimal parameters depending on the user and facial expression dynamics is out of the scope of this paper and is left for future work. Note also that the template mesh we used do not have eyeball or tongue. Modeling eye ball and tongue is left for future work. 

%

\section{Conclusion}
We proposed a method to build in real-time detailed animated 3D models of the head in dynamic scenes captured by an RGB-D camera.The contributions of this work are two fold: (1) we introduced a new 3D representation for the head by augmenting blenshape meshes with a single Deviation image, and (2) we proposed an efficient data integration technique to grow and refine our proposed 3D representation on-the-fly with input RGB-D images. The Deviation image, which augments the blendshape meshes, allowed us to capture detailed and various geometric details around the head (including the hair), while the blendshape representation allowed us to capture fine facial motions in real-time. Our proposed method do not require any training or fine fitting of the blendshape meshes to the user, which makes it easy to use and implement. We believe that our proposed method offers interesting possibilities for applications in identification and remote communication.


\ifCLASSOPTIONcaptionsoff
  \newpage
\fi




\bibliographystyle{IEEEtran}
\bibliography{IEEEabrv,paper}

\begin{thebibliography}{10}
\providecommand{\url}[1]{#1}
\csname url@samestyle\endcsname
\providecommand{\newblock}{\relax}
\providecommand{\bibinfo}[2]{#2}
\providecommand{\BIBentrySTDinterwordspacing}{\spaceskip=0pt\relax}
\providecommand{\BIBentryALTinterwordstretchfactor}{4}
\providecommand{\BIBentryALTinterwordspacing}{\spaceskip=\fontdimen2\font plus
\BIBentryALTinterwordstretchfactor\fontdimen3\font minus
  \fontdimen4\font\relax}
\providecommand{\BIBforeignlanguage}[2]{{%
\expandafter\ifx\csname l@#1\endcsname\relax
\typeout{** WARNING: IEEEtran.bst: No hyphenation pattern has been}%
\typeout{** loaded for the language `#1'. Using the pattern for}%
\typeout{** the default language instead.}%
\else
\language=\csname l@#1\endcsname
\fi
#2}}
\providecommand{\BIBdecl}{\relax}
\BIBdecl

\bibitem{Bouaziz:2013}
\BIBentryALTinterwordspacing
S.~Bouaziz, Y.~Wang, and M.~Pauly, ``Online modeling for realtime facial
  animation,'' \emph{ACM Trans. Graph.}, vol.~32, no.~4, pp. 40:1--40:10, Jul.
  2013. [Online]. Available: \url{http://doi.acm.org/10.1145/2461912.2461976}
\BIBentrySTDinterwordspacing

\bibitem{Hsieh:2015}
P.-L. Hsieh, C.~Ma, J.~Yu, and H.~Li, ``Unconstrained realtime facial
  performance capture,'' \emph{Computer Vision and Pattern Recognition (CVPR)},
  2015.

\bibitem{Weise:2011}
\BIBentryALTinterwordspacing
T.~Weise, S.~Bouaziz, H.~Li, and M.~Pauly, ``Realtime performance-based facial
  animation,'' \emph{ACM Trans. Graph.}, vol.~30, no.~4, pp. 77:1--77:10, Jul.
  2011. [Online]. Available: \url{http://doi.acm.org/10.1145/2010324.1964972}
\BIBentrySTDinterwordspacing

\bibitem{Pavan:2013}
P.~Anasosalu, D.~Thomas, and A.~Sugimoto, ``Compact and accurate 3-d face
  modeling using an rgb-d camera: Let's open the door to 3-d video
  conference,'' in \emph{Computer Vision Workshops (ICCVW), 2013 IEEE
  International Conference on}, Dec 2013, pp. 67--74.

\bibitem{Hernandez:2012}
M.~Hernandez, J.~Choi, and G.~Medioni, ``Laser scan quality 3-d face modeling
  using a low-cost depth camera,'' in \emph{Signal Processing Conference
  (EUSIPCO), 2012 Proceedings of the 20th European}, Aug 2012, pp. 1995--1999.

\bibitem{Zollhofer:2011}
M.~Zollhofer, M.~Martinek, G.~Greiner, M.~Stamminger, and J.~SuBmuth,
  ``Automatic reconstruction of personalized avatars from 3d face scans,''
  \emph{Comput. Animat. Virtual Worlds}, vol.~22, no. 2-3, pp. 195--202, 2011.

\bibitem{Newcombe:2011}
R.~Newcombe, S.~Izadi, O.~Hilliges, D.~Molyneaux, D.~Kim, A.~Davison, P.~Kohli,
  J.~Shotton, S.~Hodges, and A.~Fitzgibbon, ``Kinectfusion: Real-time dense
  surface mapping and tracking,'' \emph{Proc. of ISMAR}, pp. 127--136, 2011.

\bibitem{Newcombe:2015}
R.~A. Newcombe, D.~Fox, and S.~M. Seitz, ``Dynamicfusion: Reconstruction and
  tracking of non-rigid scenes in real-time,'' \emph{The IEEE Conference on
  Computer Vision and Pattern Recognition (CVPR)}, 2015.

\bibitem{Diego:2013}
D.~Thomas and A.~Sugimoto, ``A flexible scene representation for 3d
  reconstruction using an rgb-d camera,'' in \emph{Computer Vision (ICCV), 2013
  IEEE International Conference on}, Dec 2013, pp. 2800--2807.

\bibitem{Thomas:CVPR}
D.~Thomas and R.-i. Taniguchi, ``Augmented blendshapes for real-time
  simultaneous 3d head modeling and facial motion capture,'' in \emph{The IEEE
  Conference on Computer Vision and Pattern Recognition (CVPR)}, June 2016.

\bibitem{Cao:2013}
S.~L. C.~Cao, Y.~Weng and K.~Zhou, ``3d shape regression for real-time facial
  animation,'' \emph{ACM Transactions on Graphics}, vol.~32, no.~4, pp.
  41:1--41:10, 2013.

\bibitem{Chen:2013}
Y.-L. Chen, H.-T. Wu, F.~Shi, X.~Tong, and J.~Chai, ``Accurate and robust 3d
  facial capture using a single rgbd camera,'' in \emph{Computer Vision (ICCV),
  2013 IEEE International Conference on}, Dec 2013, pp. 3615--3622.

\bibitem{Li:2013}
\BIBentryALTinterwordspacing
H.~Li, J.~Yu, Y.~Ye, and C.~Bregler, ``Realtime facial animation with
  on-the-fly correctives,'' \emph{ACM Trans. Graph.}, vol.~32, no.~4, pp.
  42:1--42:10, Jul. 2013. [Online]. Available:
  \url{http://doi.acm.org/10.1145/2461912.2462019}
\BIBentrySTDinterwordspacing

\bibitem{Weise:2009}
\BIBentryALTinterwordspacing
T.~Weise, H.~Li, L.~Van~Gool, and M.~Pauly, ``Face/off: Live facial puppetry,''
  in \emph{Proceedings of the 2009 ACM SIGGRAPH/Eurographics Symposium on
  Computer Animation}, ser. SCA '09.\hskip 1em plus 0.5em minus 0.4em\relax New
  York, NY, USA: ACM, 2009, pp. 7--16. [Online]. Available:
  \url{http://doi.acm.org/10.1145/1599470.1599472}
\BIBentrySTDinterwordspacing

\bibitem{Cao:2014}
\BIBentryALTinterwordspacing
C.~Cao, Q.~Hou, and K.~Zhou, ``Displaced dynamic expression regression for
  real-time facial tracking and animation,'' \emph{ACM Trans. Graph.}, vol.~33,
  no.~4, pp. 43:1--43:10, Jul. 2014. [Online]. Available:
  \url{http://doi.acm.org/10.1145/2601097.2601204}
\BIBentrySTDinterwordspacing

\bibitem{Cao:2014:2}
C.~Cao, Y.~Weng, S.~Zhou, Y.~Tong, and K.~Zhou, ``Facewarehouse: A 3d facial
  expression database for visual computing,'' \emph{Visualization and Computer
  Graphics, IEEE Transactions on}, vol.~20, no.~3, pp. 413--425, March 2014.

\bibitem{Sumner:2007}
\BIBentryALTinterwordspacing
R.~W. Sumner, J.~Schmid, and M.~Pauly, ``Embedded deformation for shape
  manipulation,'' in \emph{ACM SIGGRAPH 2007 Papers}, ser. SIGGRAPH '07.\hskip
  1em plus 0.5em minus 0.4em\relax New York, NY, USA: ACM, 2007. [Online].
  Available: \url{http://doi.acm.org/10.1145/1275808.1276478}
\BIBentrySTDinterwordspacing

\bibitem{Henry:2013}
P.~Henry, D.~Fox, A.~Bhowmik, and R.~Mongia, ``Patch volumes:
  Segmentation-based consistent mapping with rgb-d cameras,'' in \emph{3D
  Vision - 3DV 2013, 2013 International Conference on}, June 2013, pp.
  398--405.

\bibitem{NieBner:2013}
\BIBentryALTinterwordspacing
M.~NieBner, M.~Zollhofer, S.~Izadi, and M.~Stamminger, ``Real-time 3d
  reconstruction at scale using voxel hashing,'' \emph{ACM Trans. Graph.},
  vol.~32, no.~6, pp. 169:1--169:11, Nov. 2013. [Online]. Available:
  \url{http://doi.acm.org/10.1145/2508363.2508374}
\BIBentrySTDinterwordspacing

\bibitem{Roth:2012}
H.~Roth and M.~Vona, ``Moving volume kinectfusion,'' \emph{British Machine
  Vision Conference (BMVC)}, 2012.

\bibitem{Whelan:2013}
T.~Whelan, M.~Kaess, J.~Leonard, and J.~McDonald, ``Deformation-based loop
  closure for large scale dense rgb-d slam,'' in \emph{Intelligent Robots and
  Systems (IROS), 2013 IEEE/RSJ International Conference on}, Nov 2013, pp.
  548--555.

\bibitem{Whelan:2012}
T.~Whelan, J.~McDonald, M.~Kaess, M.~Fallon, H.~Johannsson, and J.~J. Leonard,
  ``Kintinuous: Spatially extended kinectfusion,'' \emph{Workshop on RGB-D:
  Advanced Reasoning with Depth Cameras, in conjunction with Robotics: Science
  and Systems}, 2012.

\bibitem{Zeng:2012}
\BIBentryALTinterwordspacing
M.~Zeng, F.~Zhao, J.~Zheng, and X.~Liu, ``A memory-efficient kinectfusion using
  octree,'' in \emph{Proceedings of the First International Conference on
  Computational Visual Media}, ser. CVM'12.\hskip 1em plus 0.5em minus
  0.4em\relax Berlin, Heidelberg: Springer-Verlag, 2012, pp. 234--241.
  [Online]. Available: \url{http://dx.doi.org/10.1007/978-3-642-34263-9_30}
\BIBentrySTDinterwordspacing

\bibitem{Curless:1996}
B.~Curless and M.~Levoy, ``A volumetric method for building complex models from
  range images,'' \emph{Proc. of SIGGRAPH}, pp. 303--312, 1996.

\bibitem{Ichim:2015}
A.~E. Ichim, S.~Bouaziz, and M.~Pauly, ``Dynamic 3d avatar creation from
  hand-held video input,'' \emph{ACM Transactions on Graphics (ToG)}, 2015.

\bibitem{Garrido:2016}
\BIBentryALTinterwordspacing
P.~Garrido, M.~Zollh\"{o}fer, D.~Casas, L.~Valgaerts, K.~Varanasi,
  P.~P{\'e}rez, and C.~Theobalt, ``Reconstruction of personalized 3d face rigs
  from monocular video,'' \emph{ACM Trans. Graph.}, vol.~35, no.~3, pp.
  28:1--28:15, May 2016. [Online]. Available:
  \url{http://doi.acm.org/10.1145/2890493}
\BIBentrySTDinterwordspacing

\bibitem{Kainz:2012}
\BIBentryALTinterwordspacing
B.~Kainz, S.~Hauswiesner, G.~Reitmayr, M.~Steinberger, R.~Grasset, L.~Gruber,
  E.~Veas, D.~Kalkofen, H.~Seichter, and D.~Schmalstieg, ``Omnikinect:
  Real-time dense volumetric data acquisition and applications,'' in
  \emph{Proceedings of the 18th ACM Symposium on Virtual Reality Software and
  Technology}, ser. VRST '12.\hskip 1em plus 0.5em minus 0.4em\relax New York,
  NY, USA: ACM, 2012, pp. 25--32. [Online]. Available:
  \url{http://doi.acm.org/10.1145/2407336.2407342}
\BIBentrySTDinterwordspacing

\bibitem{Dou:2013}
M.~Dou, H.~Fuchs, and J.-M. Frahm, ``Scanning and tracking dynamic objects with
  commodity depth cameras,'' in \emph{Mixed and Augmented Reality (ISMAR), 2013
  IEEE International Symposium on}, Oct 2013, pp. 99--106.

\bibitem{Dou:2014}
M.~Dou and H.~Fuchs, ``Temporally enhanced 3d capture of room-sized dynamic
  scenes with commodity depth cameras,'' in \emph{Virtual Reality (VR), 2014
  iEEE}, March 2014, pp. 39--44.

\bibitem{Zhang:2014}
Q.~Zhang, B.~Fu, M.~Ye, and R.~Yang, ``Quality dynamic human body modeling
  using a single low-cost depth camera,'' in \emph{Computer Vision and Pattern
  Recognition (CVPR), 2014 IEEE Conference on}, June 2014, pp. 676--683.

\bibitem{bodyfusion}
T.~Yu12, K.~Guo, F.~Xu, Y.~Dong, Z.~Su, J.~Zhao, J.~Li, Q.~Dai, and Y.~Liu,
  ``Bodyfusion: Real-time capture of human motion and surface geometry using a
  single depth camera,'' \emph{IEEE International Conference on Computer Vision
  (ICCV)}, 2017.

\bibitem{volumedeform}
M.~Innmann, M.~Zollh{\"o}fer, M.~Nie{\ss}ner, C.~Theobalt, and M.~Stamminger,
  ``Volumedeform: Real-time volumetric non-rigid reconstruction,'' in
  \emph{European Conference on Computer Vision}.\hskip 1em plus 0.5em minus
  0.4em\relax Springer, 2016, pp. 362--379.

\bibitem{killingfusion}
M.~Slavcheva, M.~Baust, D.~Cremers, and S.~Ilic, ``Killingfusion: Non-rigid 3d
  reconstruction without correspondences,'' \emph{IEEE Conference on Computer
  Vision and Pattern Recognition (CVPR)}, vol.~3, no.~4, p.~7, 2017.

\bibitem{thies2015}
J.~Thies, M.~Zollh{\"o}fer, M.~Nie{\ss}ne, L.~Valgaerts, M.~Stamminger, and
  C.~Theobalt, ``Real-time expression transfer for facial reenactment,''
  \emph{ACM Transactions on Graphics (ToG)}, vol.~34, no.~6, pp. 183--1, 2015.

\bibitem{cao2015real}
C.~Cao, D.~Bradley, K.~Zhou, and T.~Beeler, ``Real-time high-fidelity facial
  performance capture,'' \emph{ACM Transactions on Graphics (ToG)}, vol.~34,
  no.~4, p.~46, 2015.

\bibitem{Cao2016}
C.~Cao, H.~Wu, Y.~Weng, T.~Shao, and K.~Zhou, ``Real-time facial animation with
  image-based dynamic avatars,'' \emph{ACM Transactions on Graphics (ToG)},
  vol.~35, no.~4, 2016.

\bibitem{Torre:2015}
F.~De~la Torre, W.-S. Chu, X.~Xiong, F.~Vicente, X.~Ding, and J.~Cohn,
  ``Intraface,'' in \emph{Automatic Face and Gesture Recognition (FG), 2015
  11th IEEE International Conference and Workshops on}, vol.~1, May 2015, pp.
  1--8.

\bibitem{Zhou:2013}
Q.-Y. Zhou, S.~Miller, and V.~Koltun, ``Elastic fragments for dense scene
  reconstruction,'' in \emph{Computer Vision (ICCV), 2013 IEEE International
  Conference on}, Dec 2013, pp. 473--480.

\bibitem{Sumner:2004}
\BIBentryALTinterwordspacing
R.~W. Sumner and J.~Popovi\'{c}, ``Deformation transfer for triangle meshes,''
  \emph{ACM Trans. Graph.}, vol.~23, no.~3, pp. 399--405, Aug. 2004. [Online].
  Available: \url{http://doi.acm.org/10.1145/1015706.1015736}
\BIBentrySTDinterwordspacing

\bibitem{Rusinkiewicz:2001}
S.~Rusinkiewicz and M.~Levoy, ``Efficient variants of the icp algorithm,'' in
  \emph{3-D Digital Imaging and Modeling, 2001. Proceedings. Third
  International Conference on}, 2001, pp. 145--152.

\bibitem{Lie}
J.-L. Blanco, ``A tutorial on se (3) transformation parameterizations and
  on-manifold optimization,'' \emph{Technical report, Universidad de Malaga},
  2014.

\bibitem{Sugimoto:1995}
T.~Sugimoto, M.~Fukushima, and T.~Ibaraki, ``A parallel relaxation method for
  quadratic programming problems with interval constraints,'' \emph{Journal of
  Computational and Applied Mathematics}, vol. 60(12), pp. 219--236, June 1995.

\bibitem{SLIC}
R.~Achanta, A.~Shaji, K.~Smith, A.~Lucchi, P.~Fua, and S.~Susstrunk, ``Slic
  superpixels compared to state-of-the-art superpixel methods,'' \emph{IEEE
  Trans. on PAMI}, vol. 34(11), pp. 2274--2282, November 2012.

\end{thebibliography}

\begin{IEEEbiography}[{\includegraphics[width=1in,height=1.25in,clip,keepaspectratio]{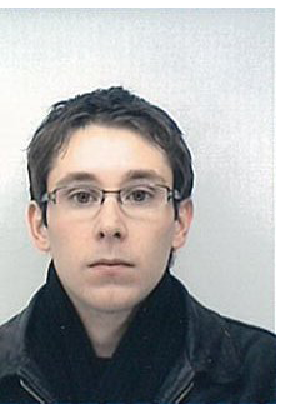}}]{Diego Thomas}
received his master's degree in Informatics and Applied Mathematics from the Ecole Nationale Superieure d'Informatique et de Mathematiques Appliquees de Grenoble (ENSIMAG), France in 2008. He received his Ph.D. from the Graduate University for Advanced Studies (SOKENDAI), Japan in 2012. His research interests are 3D images registration, 3D reconstruction and photometric analysis.
\end{IEEEbiography}







\end{document}